\pdfoutput=1
\PassOptionsToPackage{table,xcdraw}{xcolor} 
\documentclass[11pt]{article}

\usepackage[final]{acl}

\usepackage{times}
\usepackage{latexsym}

\usepackage[T1]{fontenc}

\usepackage[utf8]{inputenc}

\usepackage{microtype}

\usepackage{inconsolata}

\usepackage{graphicx}
\usepackage{amsmath}
\usepackage{csquotes}
\usepackage{booktabs}
\usepackage[most]{tcolorbox}
\usepackage{algorithm, algorithmic}
\usepackage{amssymb}
\usepackage{amsmath}
\usepackage{multirow}
\usepackage{lipsum}
\usepackage{tabularx}
\usepackage{bm}
\usepackage{diagbox}
\usepackage[normalem]{ulem}
\useunder{\uline}{\ul}{}
\usepackage{float} 
\usepackage{pifont} 
\usepackage{array}
\usepackage{subfig}
\usepackage{fontawesome}

\newcommand{\cmark}{\ding{51}} 
\newcommand{\xmark}{\ding{55}} 
\title{Towards Explainable Temporal Reasoning in Large Language Models: A Structure-Aware Generative Framework}

\newcommand*{\affaddr}[1]{#1} 
\newcommand*{\affmark}[1][*]{\textsuperscript{#1}} 
\newcommand*{\email}[1]{\texttt{#1}} 

\author{
Zihao Jiang\affmark[1]$^{*}$,
Ben Liu\affmark[1]\thanks{~~Equal contribution.},
Miao Peng\affmark[2],
Wenjie Xu\affmark[1],
Yao Xiao\affmark[1],
Zhenyan Shan\affmark[1],
Min Peng\affmark[1]\thanks{~~Corresponding author} \\
\affaddr{\affmark[1]School of Computer Science, Wuhan University, China}\\
\affaddr{\affmark[2]The Hong Kong University of Science and Technology (Guangzhou)}\\
\email{\{jiangzihao,liuben123,vingerxu,y.xiao,bbcavendish,pengm\}@whu.edu.cn}\\
\email{mpeng885@connect.hkust-gz.edu.cn}
}

\begin{document}
\maketitle
\begin{abstract}
While large language models (LLMs) show great potential in temporal reasoning, most existing work focuses heavily on enhancing performance, often neglecting the explainable reasoning processes underlying the results. To address this gap, we introduce a comprehensive benchmark covering a wide range of temporal granularities, designed to systematically evaluate LLMs' capabilities in explainable temporal reasoning. Furthermore, our findings reveal that LLMs struggle to deliver convincing explanations when relying solely on textual information. To address challenge, we propose \textbf{GETER}, a novel structure-aware generative framework that integrates \textbf{G}raph structures with text for \textbf{E}xplainable \textbf{TE}mporal \textbf{R}easoning. Specifically, we first leverage temporal knowledge graphs  to develop a temporal encoder that captures structural information for the query. Subsequently, we introduce a structure-text prefix adapter to map graph structure features into the text embedding space. Finally, LLMs generate explanation text by seamlessly integrating the soft graph token with instruction-tuning prompt tokens. Experimental results indicate that GETER achieves state-of-the-art performance while also demonstrating its effectiveness as well as strong generalization capabilities. Our dataset and code are available at \url{https://github.com/carryTatum/GETER}.
\end{abstract}

\section{Introduction}

Temporal reasoning (TR) is a fundamental cognitive skill essential for understanding complex tasks like planning and causal relation discovery~\cite{relatedwork_3}. In natural language processing (NLP), temporal reasoning refers to a model's capability to effectively comprehend, represent, and predict time-sensitive contexts~\cite{DBLP:conf/emnlp/YangLFC24}. This capability is critical for real-world applications that depend on temporal data, including search engine recommendations~\cite{DBLP:journals/umuai/BoginaKJBKT23} and news article aggregation ~\cite{DBLP:journals/corr/abs-2501-00888}. 

Recently, large language models (LLMs) have demonstrated remarkable performance in tackling complex tasks~\cite{DBLP:journals/tmlr/WeiTBRZBYBZMCHVLDF22,DBLP:conf/acl/0009C23,GPT-4o,DBLP:journals/corr/abs-2503-00845,DBLP:conf/www/LiuZLYPY25}. Building on this success, recent studies have increasingly focused on exploring the TR capabilities of LLMs. These works primarily adopt general approaches to evaluate and enhance the TR capabilities of LLMs. For instance, ~\citet{relatedwork_2} and ~\citet{relatedwork_4} design time-sensitive queries to benchmark LLMs, while~\citet{DBLP:conf/acl/Wang024} and~\citet{DBLP:conf/acl/ChuCCY00024} extend these efforts by using prompting strategies like in-context learning (ICL) and Chain-of-Thought (CoT) reasoning for comprehensive evaluation. Furthermore, ~\citet{ICL} and~\citet{COH} employ ICL with prompts containing intermediate reasoning steps to guide models, while ~\citet{GenTKG} and ~\citet{COH2} adopt fine-tuning methods, training LLMs on reasoning process texts to enable them to produce accurate answers.


\begin{figure}[!t]
    \centering
    \includegraphics[width=\linewidth]{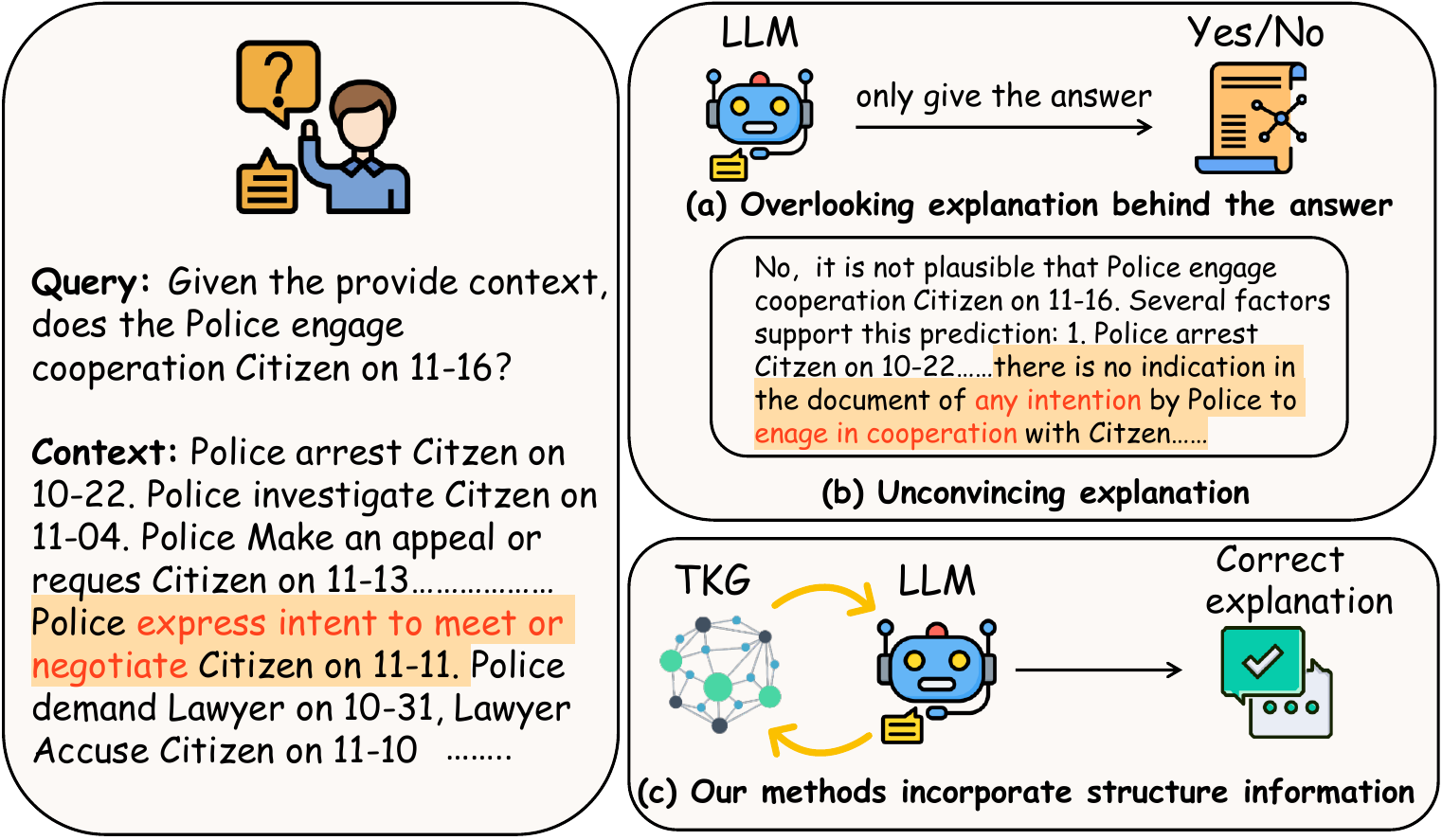}
    \caption{An illustration of existing temporal reasoning works highlights the lack of focus on explanations behind the reasoning. Meanwhile, LLMs often struggle to generate convincing answers due to hallucinations.}
    \label{fig:introduction}
\end{figure}


Although existing methods have explored LLMs' potential in temporal reasoning, they exceedingly focus on improving performance, often overlooking the explainable reasoning processes behind the results, as illustrated in Figure~\ref{fig:introduction}(a). The study of explainable temporal reasoning is crucial, as it promotes transparency, enhances effectiveness, and fosters trust in understanding temporal dynamics. Moreover, with their impressive semantic understanding and generation capabilities, LLMs are uniquely positioned to address the challenges of explainable reasoning~\cite{DBLP:conf/iclr/0002WSLCNCZ23,DBLP:conf/emnlp/MaR024}, as they can generate flexible, human-readable reasoning processes. Therefore, we posit the following research question to guide our study: \textit{Can LLMs effectively make accurate predictions and clearly explaining their reasoning processes in complex temporal reasoning scenarios?}

To address this challenge, we propose the \textbf{ETR} benchmark, a comprehensive benchmark for explainable temporal reasoning. Specifically, ETR consists of five datasets covering a wide range of temporal granularities \textbf{(minutes, days, and years)}. Each instance is represented as a triple of \textit{<query text, reasoning chains text, explanation text>} where the query and related reasoning chains are derived from Temporal Knowledge Graphs (TKGs). The explanation text is synthesized using GPT-4o~\cite{GPT-4o} with constrained generation prompt protocols, taking the query and reasoning chains as input. The resulting explanation text effectively integrates both the original gold prediction and the underlying reasoning processes. ETR aims to challenge LLMs not only to predict future events from the given reasoning chains text but also to generate explanations of their reasoning processes.

Building on this benchmark, we identify that the key to achieving explainable temporal reasoning lies in enabling LLMs to capture structured patterns that reflect the relationships and dynamics between events over time. As shown in Figure~\ref{fig:introduction}(b), our findings reveal that LLMs struggle to deliver convincing explanations when relying solely on textual information, a challenge (e.g. hallucinations) also highlighted in previous analyses~\cite{G-Retriever,Filter-then-Generate}. To address this challenge, we propose a novel structure-aware generative framework \textbf{GETER}, which advances explainable temporal reasoning by effectively bridging the gap between graph structures and text. Specifically, we leverage TKGs to develop a temporal encoder that captures structural information. Subsequently, the encoder converts the query and reasoning chains into a soft graph token, which is then mapped into the LLM's text space via a lightweight adapter. Finally, LLM can generate explanation text by integrating the soft graph token with instruction-tuning prompt tokens, seamlessly combining structural and contextual semantic information. Experimental results show that our proposed GETER achieves state-of-the-art performance. In summary, the contributions of this paper are as follows:
\begin{itemize}
    \item We introduce ETR, a comprehensive benchmark covering a wide range of temporal granularities for systematically evaluating LLMs' explainable temporal reasoning.
    
    \item To bridge the gap between graph structures and text, we propose GETER, a novel structure-aware generative framework which leverages a lightweight structure-text adapter to enhance LLMs' explainable temporal reasoning capabilities.

    \item Our GETER achieves state-of-the-art performance on five datasets using widely-used LLMs, demonstrating the superiority of our model. Further experiments reveal the effectiveness and strong generalization ability of GETER.

\end{itemize} 

\section{Related Work}

\subsection{LLMs for Temporal Reasoning}
With the rapid advancement of LLMs, research has increasingly focused on evaluating and enhancing their temporal reasoning capabilities. Existing studies primarily leverage the parametric knowledge of LLMs to assess and improve performance. For instance, several studies~\cite{relatedwork_2,relatedwork_4} design time-sensitive queries to benchmark LLMs, while others~\cite{DBLP:conf/acl/Wang024,DBLP:conf/acl/ChuCCY00024} extend these efforts to diverse temporal reasoning tasks using general evaluation methods. Additionally, some methods~\cite{ICL,COH} utilize in-context learning by providing prompts with demonstrations of intermediate reasoning steps to guide the model, whereas fine-tuning methods~\cite{GenTKG,COH2} train LLMs on reasoning texts to enable them to generate accurate final answers. Despite these advancements, most efforts focus on improving performance through parametric knowledge, with limited emphasis on explanation.

\subsection{Explainable Temporal Reasoning}
In temporal reasoning tasks, explainability is crucial for ensuring transparency, trust, and reliability. Existing works for explainable temporal reasoning primary fall into two categories: logic rule-based methods and reinforcement learning-based methods. Logic rule-based methods~\cite{Tlogic,TECHS,DBLP:conf/emnlp/MeiYCJ22} ensure explainability through explicit rule templates but struggle to balance generalization and explainability in complex scenarios. Reinforcement learning-based methods~\cite{xERTE,TITer} construct reasoning paths guided by predefined reward mechanisms. However, their explainability is limited by the implicit nature of their decision-making processes. In contrast, LLMs offer unique advantages for explainable reasoning by leveraging semantic understanding and generation capabilities~\cite{relatedwork_2,relatedwork_7}, enabling more flexible and human-readable reasoning processes.  While~\citet{Back_to_future} conduct a preliminary exploration of LLM explainability, their work overlooks finer-grained temporal dimensions evaluation and fails to enhance LLMs through the integration of temporal graph features.

\section{Proposed ETR Benchmark}

\subsection{Problem Definition} 

Temporal Knowledge Graphs (TKGs) $\mathcal{G}$ are represented as a sequence of KGs $(\mathcal{G}_0, \mathcal{G}_1, \ldots, \mathcal{G}_t)$ arranged by timestamp $t$. Let $\mathcal{G}= (\mathcal{E},\mathcal{R},\mathcal{F})$ be a TKG instance, where $\mathcal{E}$,$\mathcal{R}$, $\mathcal{F}$ represent the set of entities, relations and facts, respectively. Each fact can be represented as a quadruple $(e_s, r, e_o, t) \in \mathcal{F}$, where subject and object $e_s, e_o \in \mathcal{E}$, relation $r \in \mathcal{R}$. Explainable temporal reasoning aims to challenge LLMs to predict future events based on reasoning chains and generate explanations of their reasoning. Formally, given reasoning chains $\mathcal{C}$ consisting of facts $\mathcal{F}_{[t_q - w, t_q)}$, the task is to predict the probability that a query $q$ will occur at future time $t_q$, where $w$ is the window size. Based on this probability, the model classifies $q$ into one of three categories: "Yes", "No", or "Unsure", and generates an explanation for its prediction. The prediction and explanation together form the final output $A$. To train and evaluate the model, we define two types of instances: training instances $\mathcal{T}_{train}$ and test instances $\mathcal{T}_{test}$. These instances follow the extrapolation condition~\cite{RE-NET}, where the training time ($t_{train}$) strictly precedes the test time ($t_{test}$), i.e., $t_{train} < t_{test}$. Each instance $\mathcal{T}_i$ consists of the following components: the query text $\mathcal{Q}_i$, the input reasoning chains text $\mathcal{C}_i$, and explanation text $\mathcal{A}_i$, formally defined as:$\mathcal{T}_i = \{Q_i, \mathcal{C}_i, \mathcal{A}_i\}$. 
\subsection{Pipeline}
\begin{figure*}[!t]
    \centering
    \includegraphics[width=0.95\textwidth]{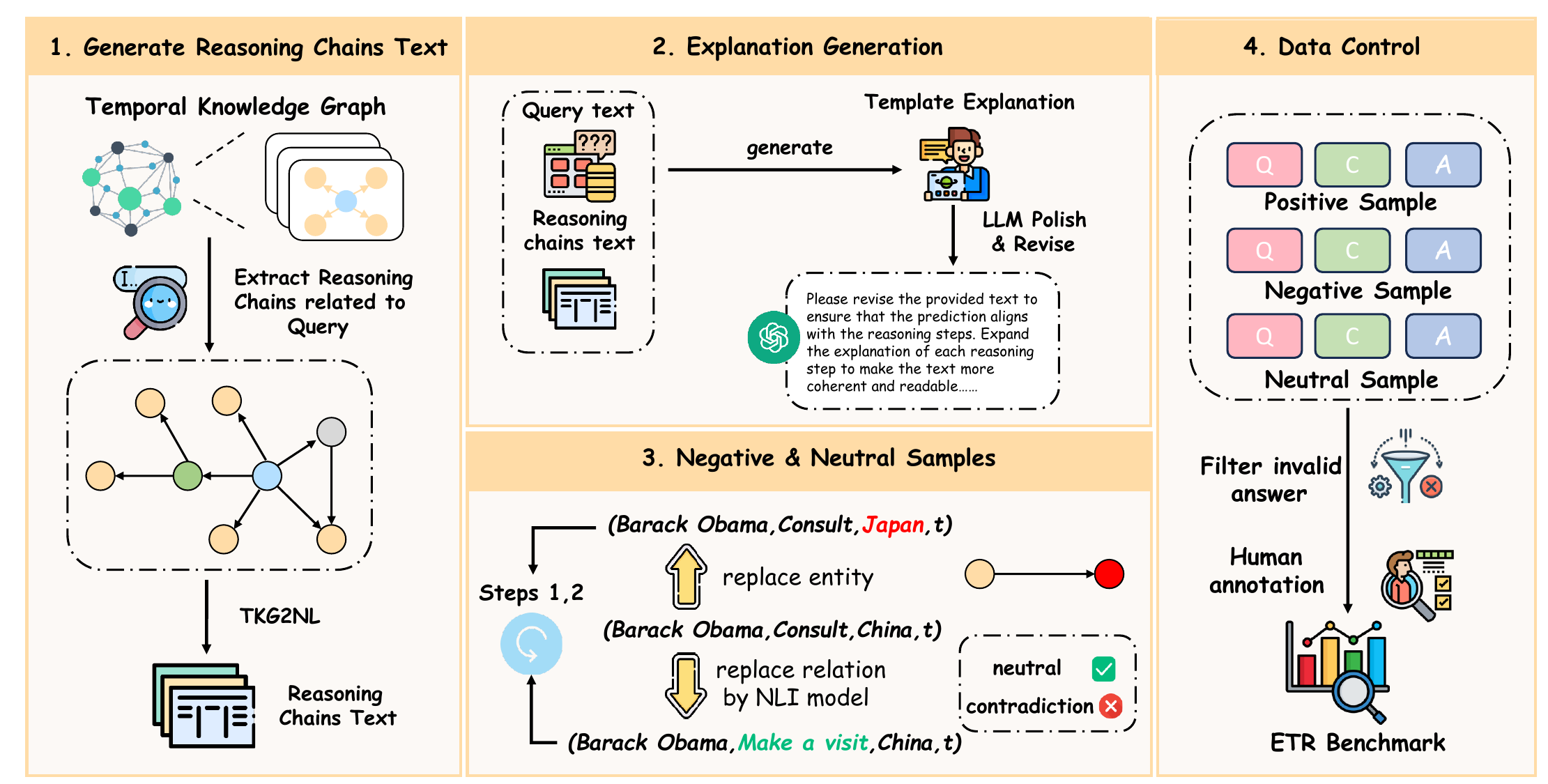}
    \caption{The pipeline of generating \textbf{ETR} benchmark.}
    \label{fig:Datasets}
\end{figure*}
As illustrated in Figure~\ref{fig:Datasets}, we present \textbf{ETR}, a comprehensive benchmark for \textbf{E}xplainable \textbf{T}emporal \textbf{R}easoning.  To accomplish this goal, we extract reasoning chains for each query and generate explanation text using GPT-4o. Additionally, we sample negative and neutral examples in a similar manner to provide a thorough evaluation of the LLMs. The detailed construction process is outlined as follows.

\subsubsection{Reasoning Chains Text Construction}
To construct reasoning chains text, given a query $q = (e_s,r,e_o,t_q)$, we extract the graph reasoning chains $\mathcal{C}(e_s,e_o)$ associated with entities $e_s$ and $e_o$ using a breadth-first search (BFS) methods~\cite{DBLP:conf/emnlp/JiangZZLW23}. The extraction process considers reasoning chains occurring within the time interval $[t_q - w, t_q)$ and is formalized as follows:
\begin{equation}
\label{equation:1}
\begin{split}
\mathcal{C}(e_s,e_o) \leftarrow \bigwedge_{i=1}^{l} (E_i, R_i, E_{i+1}, T_i),
\end{split}
\end{equation}
where $E_1=e_s$, $E_{l+1} = e_o$, $l \in \{1, 2\}$ denotes the path length. Here, $E_i$ represents the entity, $R_i$ denotes the relation, and $T_i$ is the corresponding timestamp. Once these reasoning chains $\mathcal{C}(e_s, e_o)$ are extracted, they are converted into natural language sentences to form the input text $\mathcal{C}_i$. 

\subsubsection{Explanation Generation}
Based on the query $q = (e_s,r,e_o,t_q)$ and reasoning chains $\mathcal{C}(e_s,e_o)$, we employ a template to generate an initial explanation text $\mathcal{A}_i ^{'}$ as follows:

\begin{tcolorbox}
    [colback=gray!20, colframe=gray!100, sharp corners, leftrule={4pt}, rightrule={0pt}, toprule={0pt}, bottomrule={0pt}, left={8pt}, right={10pt}, top={2pt}, bottom={2pt}, halign=center]
    \begin{displayquote}
    \small
    \emph{We predict that [$e_s$] [$r$] [$e_o$] will happen on [$t_q$]. Here are the reasoning steps: $\mathcal{C}(e_s,e_o)$.}
    \end{displayquote}
\end{tcolorbox}

However, not all reasoning chains can adequately justify the occurrence of the given query, and the template-generated explanation text often exhibits issues such as incoherence, unnatural flow, and insufficient logical consistency, ultimately failing to provide a clear and compelling rationale. To address these limitations, we employ GPT-4o to enhance the quality of the final explanations $\mathcal{A}_i$, guided by the prompt provided in Appendix~\ref{sec:Prompt for Generating Explanations of Positive Samples}

\subsubsection{Negative and Neutral samples}
To evaluate the ability of LLMs in explainable temporal reasoning, particularly in inferring logical correlations between the queries and historical facts, we introduce negative and neutral samples. Negative samples are used to test the model's ability to reject logically inconsistent or counterfactual scenarios, while neutral samples assess its capacity to infer uncertainty and ambiguity in scenarios with insufficient evidence.

\begin{table}[!t]
\renewcommand{\arraystretch}{1.2}
\centering
\resizebox{\linewidth}{!}{%
\begin{tabular}{c|c|c|ccc|c}
\toprule
\textbf{Dataset} & \textbf{Time Granularity} & \textbf{Type} & $|Pos.|$ & $|Neg.|$ & $|Neu.|$ & \textbf{Total} \\ \hline
\midrule
\multirow{2}{*}{ICEWS14} & \multirow{2}{*}{1 day} & \textit{Train} & 5000 & 4800 & 4500 & 14300 \\ \cline{3-7}
                         &                        & \textit{Test}  & 800  & 700  & 600  & 2100  \\ \hline
\multirow{2}{*}{ICEWS05-15} & \multirow{2}{*}{1 day} & \textit{Train} & 4500 & 4400 & 4200 & 13100 \\ \cline{3-7}
                         &                        & \textit{Test}  & 720  & 680   & 660   & 2060   \\ \hline
\multirow{2}{*}{ICEWS18} & \multirow{2}{*}{1 day} & \textit{Train} & 4400 & 4200 & 4000 & 12600 \\ \cline{3-7}
                            &                        & \textit{Test}  & 750 & 700 & 650 & 2100 \\ \hline
\multirow{2}{*}{GDELT} & \multirow{2}{*}{15 minutes} & \textit{Train} & 4800 & 4600 & 4400 & 13800 \\ \cline{3-7}
                       &                          & \textit{Test}  & 800 & 700 & 650 & 2150 \\ \hline
\multirow{2}{*}{WIKI} & \multirow{2}{*}{1 year} & \textit{Train} & 2482 & 2504 & 2342 & 7328 \\ \cline{3-7}
                       &                          & \textit{Test}  & 347 & 286 & 316 & 949 \\
\hline
\bottomrule
\end{tabular}%
}
\caption{Statistics of the \textbf{ETR} benchmark. $|Pos.|$, $|Neg.|$, and $|Neu.|$ denote the number of positive, negative, and neutral samples, respectively.}
\label{tab: sampleDatasets}
\end{table}

\textbf{Negative Samples.} Negative samples represent counterfactual queries. To achieve this goal, we modify the positive query quadruple $q = (e_s, r, e_o, t_q)$ by replacing $o$ with a different entity $o'$, resulting in $q' = (e_s, r, e_o', t_q)$, where $q' \notin \mathcal{F}$. This creates a hard negative sample that introduces factual inconsistencies. Additionally, we derive negative sample reasoning chains $\mathcal{C}(e_s, e_o')$ as defined in Equation~\ref{equation:1}. Following a similar process for positive samples, we design the corresponding prompt for GPT-4o, detailed in Appendix~\ref{sec:Prompt for Generating Explanations of Negative Samples}.

\textbf{Neutral Samples.} In neutral samples, LLMs are expected to predict "unsure" for the query, as the reasoning chain lacks sufficient evidence to support or refute it. To construct these samples, we replace the positive query relation $q = (e_s, r, e_o, t_q)$ with $q'' = (e_s, r', e_o, t_q)$, where $r'$ is a semantically \textit{neutral relation} to $r$ and $q'' \notin \mathcal{F}$. The neutral relation $r'$ is identified using a Natural Language Inference (NLI) model~\cite{NLI}, which classifies relationships into entailment, contradiction, and neutral. We select $r'$ as neutral only if the NLI model assigns $P(\text{neutral}) > \tau$, where $\tau$ is a predefined threshold. The reasoning chains for neutral samples, $\mathcal{C}(e_s, e_o)$, are consistent with those of positive samples. Details of the GPT-4o prompt are provided in Appendix~\ref{sec:Prompt for Generating Explanations of Neutral Samples}.

\subsection{Benchmark Summary and Evaluation}

As summarized in Table~\ref{tab: sampleDatasets}, the proposed benchmark covers a wide range of temporal granularities.  To achieve this goal, we use five widely adopted temporal knowledge graph reasoning datasets: ICEWS14~\cite{DBLP:conf/emnlp/Garcia-DuranDN18}, ICEWS18~\cite{xERTE}, ICEWS05-15~\cite{DBLP:conf/emnlp/Garcia-DuranDN18}), GDELT~\cite{GenTKG}, and WIKI~\cite{DBLP:conf/www/LeblayC18}. To ensure the quality of the dataset, we filter out invalid answers and conduct human evaluation. Further details refer to Appendix~\ref{sec:sec:Benchmark Summary and Evaluation}.

\begin{figure*}[!t]
    \centering
    \includegraphics[width=0.9\textwidth]{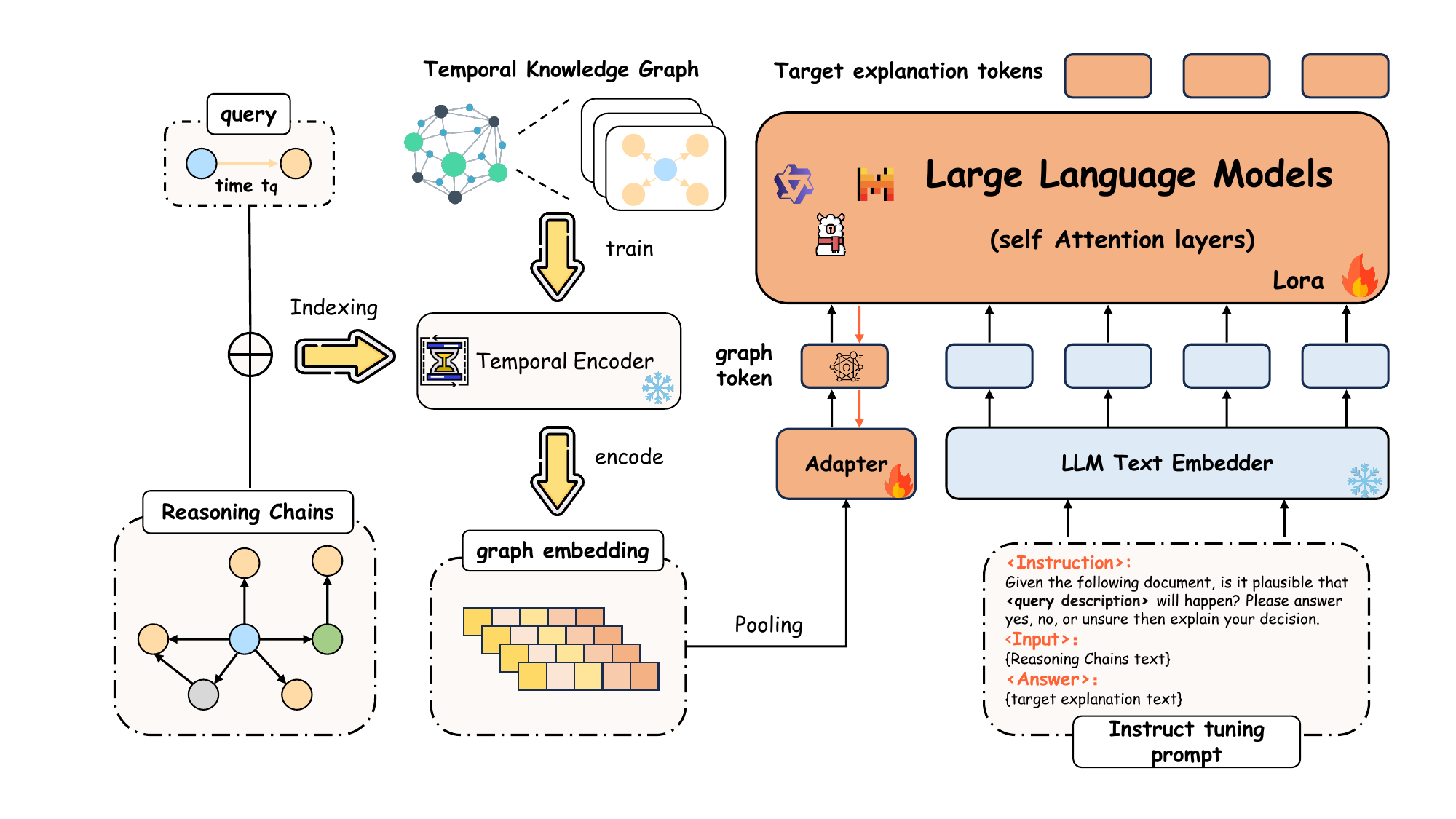}
   \caption{The overall framework of \textbf{GETER}. To bridge the gap between graph and text, we leverage TKGs to train a temporal encoder that captures structural information. Subsequently, the query and reasoning chains are encoded into a soft graph token, which is mapped into the text embedding space through a lightweight adapter. Finally, the target explanation text is generated using the soft graph token and related instruction tuning prompt tokens.}
    \label{fig:Model}
\end{figure*}

\section{Methodology}

In this section, we present \textbf{GETER}, a novel structure-aware generative framework that integrates \textbf{G}raph structures with text for \textbf{E}xplainable \textbf{TE}mporal \textbf{R}easoning. The overall architecture of our proposed model is illustrated in Figure~\ref{fig:Model}. Specifically, we first leverage a temporal encoder to obtain structural embeddings for both entities and relations. Subsequently, we introduce a structure-text prefix adapter as described in Sec.~\ref{sec:Structure-Text Adapter} to map graph structure features into the text embedding space. Finally, we apply an instruction-tuning strategy (Sec.~\ref{sec:Instruction Tuning Strategy}) to effectively adapt the model to the explainable temporal reasoning task.

\subsection{Indexing}
\label{sec:Indexing}


We aim to harness the semantic understanding and temporal reasoning capabilities of LLMs for the explainable temporal reasoning task. However, relying solely on LLMs within a text-based prediction framework to infer correlations between queries and reasoning chains inevitably neglects the structural information in the TKG $\mathcal{G}$. To address this, we first employ a temporal encoder (TKG model), such as RE-GCN~\cite{RE-GCN}, which utilizes the message-passing mechanism of GNNs to effectively capture structural patterns, to generate the structural representation $\mathbf{s_n}$:
\begin{equation}
\mathbf{s_n} = \textit{TemporalEncoder}(x_n | \mathcal{G}) \in \mathbb{R}^{d_{s}},
\end{equation}
where $x_n$ represents the initialized embedding of entity or relation $n$, and $d_{s}$ denotes the dimension of the structural embedding. In this way, we get entity embedding matrix $\mathbf{E} \in \mathbb{R}^{|\mathcal{E}| \times d_s}$ and relation embedding matrix $\mathbf{R} \in \mathbb{R}^{|\mathcal{R}| \times d_s}$, respectively.

\subsection{Structure-Text Adapter}
\label{sec:Structure-Text Adapter}

To effectively integrate structure-based embeddings of entities and relations with textual information, we propose a soft prompt strategy that combines structural and textual features in a contextualized manner. Specifically, given the query $q=(e_s,r,e_o,t)$ and reasoning chains $\mathcal{C}(e_s, e_o)$, we compute the representation of the query and reasoning chains via parameter-free message passing on the encoded structural features. The resulting graph representation is then projected into the embedding space of LLMs using a trainable projection matrix $\mathbf{W}_p \in \mathbb{R}^{3d_s \times d_x}$, as follows:
\begin{align}
\mathbf{S}_{\mathcal{C}(e_s, e_o)} &=  \sum_{(e_s', r',e_o')\in \mathcal{C}(e_s, e_o)} (\mathbf{e_s'} \Vert \mathbf{r'} \Vert \mathbf{e_o'}) , \\
\mathbf{S}_{graph} &= \mathbf{W}_p \cdot \frac{\mathbf{S}_{\mathcal{C}(e_s, e_o)} + \mathbf{S}_{q}}{|\mathcal{C}(e_s, e_o)|+1},
\end{align}
where $\Vert$ denotes concatenation, $\mathbf{S}_{q} = (\mathbf{e_s} \Vert \mathbf{r} \Vert \mathbf{e_o})$, $\mathbf{S}_{graph}$ is the projected graph representation, and $d_x$ denotes the dimension of embedding space of LLMs. $\mathbf{e_s'} \in \mathbb{R}^{1 \times d_s}$, $\mathbf{r'} \in \mathbb{R}^{1 \times d_s}$, and $\mathbf{e_o'} \in \mathbb{R}^{1 \times d_s}$ are the embeddings of the subject entity, relation, and object entity, respectively. This straightforward linear mapping is adopted due to its proven effectiveness in aligning graph-based and textual representations~\cite{G-Retriever,Filter-then-Generate}.

\subsection{Instruction Tuning Strategy}
\label{sec:Instruction Tuning Strategy}

The instruction tuning process is designed to adapt the reasoning behavior of the LLM to align with the specific constraints and requirements of the explainable temporal reasoning task. To facilitate the generation of the target explainable text, we provide the corresponding query text $\mathcal{Q}$ and reasoning chains text $\mathcal{C}(e_s, e_o)$ as inputs to the LLM, which produce their textual representations, denoted as $X = X_{\mathcal{Q}} + X_{\mathcal{C}}$. Let $\bm{X} \in \mathbb{R}^{|X| \times d_x}$ represent the textual content embeddings of the input, where $|X|$ denotes the token length of $X$. The final input to the LLM is constructed by concatenating the soft graph token embeddings $\mathbf{S}_{graph}$ (as described in Sec.~\ref{sec:Structure-Text Adapter}) with the textual embedding, expressed as $\bm{X}' = \mathbf{S}_{graph} \Vert \bm{X}$. Lastly, our optimization objective is to maximize the likelihood of generating the target explanation text $Y_{\mathcal{A}}$:
\begin{equation}
P(\bm{Y}_{\mathcal{A}} | \bm{X}', \bm{X}_{\mathcal{I}}) = \prod_{j=1}^{L} P_\theta \big( y_j \big| \bm{X}', \bm{X}_{\mathcal{I}}, \bm{Y}_{<j} \big),
\end{equation}
where $\bm{X}_{\mathcal{I}}$ denotes the representation of instruction tokens, $L$ is the token length of the target explanation text, and $Y_{<j}$ represents the prefix of the missing explanation text sequence $Y_{\mathcal{A}}$ up to position $j - 1$. Considering the overhead of updating all parameters in LLMs, we adopt Low-Rank Adaptation (LoRA) technique~\cite{LORA} for its effectiveness~\cite{DBLP:conf/nips/LiuTMMHBR22}. The example of instruction data can be seen in Appendix~\ref{sec:Example Prompt for Instruction Tuning}.


\section{Experiments}

\subsection{Experiments Setup}
\label{Experiments Setup}
\noindent \textbf{Baselines.} We evaluate our benchmark with four representative LLMs: GPT-4o~\cite{GPT-4o}, Llama3-8B-Instruct~\cite{Llama3-8b}, Qwen2.5-7B-Instruct~\cite{qwen2}, and Mistral-7B-Instruct-v0.3~\cite{mistral}. For our framework, we adopt open-source LLMs as backbones and use RE-GCN~\cite{RE-GCN} as temporal encoder. Implementation details refer to Appendix~\ref{sec:Implementation Details}. Furthermore, performance comparisons with four graph-based methods (RE-GCN, CEN~\cite{CEN}, CENET~\cite{CENET}, and SiMFy~\cite{SiMFy}) are presented in Appendix~\ref{Comparison with Graph-based Methods}.

\noindent \textbf{Metrics.} We evaluate explainable temporal reasoning capabilities of models in two aspects: prediction and explanation. For prediction, we report precision, recall, and F1 scores. For explanation, we employ BLEU~\cite{BLEU} (4-gram), ROUGE~\cite{ROUGE} (ROUGE-L), METEOR~\cite{METEOR}, and BertScore~\cite{BERTScore} to measure the similarity between model-generated explanations and the ground truth in the test set. 

\begin{table*}[htb]
\begin{center}
\resizebox{1.0\textwidth}{!}{
\begin{tabular}{c|c|cccc|cccc|cccc} 
\toprule
\multirow{2}{*}{Models} & \multirow{2}{*}{\diagbox{Types}{Datasets}} & \multicolumn{4}{c|}{ICEWS14} & \multicolumn{4}{c|}{GDELT} & \multicolumn{4}{c}{ICEWS05-15} \\ \cline{3-14}
 &  & Positive & Negative & Neutral & Overall & Positive & Negative & Neutral & Overall & Positive & Negative & Neutral & Overall \\ \midrule
 
 
\multirow{2}{*}{GPT-4o} 
 & zero-shot\textit{\ w/o chains text} & 53.13 & 20.02 & 12.95 & 30.61 & 19.08 & 43.78 & 25.50 & 29.06 & 55.45 & 26.33 & 15.47 & 33.03 \\
 & zero-shot & 60.10 & 9.54 & 48.56 & 39.95 & 42.74 & 37.16 & 29.21 & 36.83 & 61.63 & 11.89 & 47.16 & 40.58 \\ \midrule
\multirow{4}{*}{Llama3-8B-Instruct} 
 & zero-shot\textit{\ w/o chains text} & 21.69 & 27.11 & 35.42 & 27.42 & 1.95 & 33.13 & 39.44 & 23.44 & 11.75 & 28.98 & 39.41 & 26.30 \\
 & zero-shot & 56.51 & 10.20 & 6.20 & 26.70 & 53.48 & 15.62 & 29.47 & 33.90 & 57.14 & 17.50 & 14.03 & 30.24 \\
 & LoRA \textit{\ w/o chains text} & 62.27 & 36.98 & 48.17 & 49.81 & 61.94 & 7.19 & 69.14 & 46.29 & 65.67 & 38.56 & 68.02 & 57.47 \\
 & LoRA & 70.37 & 58.06 & 67.99 & 65.59 & \textbf{62.86} & 28.57 & 78.56 & 56.44 & 71.32 & 51.77 & 74.40 & 65.86 \\ 
 & \textbf{GETER} & \textbf{75.07} & \textbf{67.38} & \textbf{81.15} & \textbf{74.25} & 62.62 & \textbf{68.74} & \textbf{88.73} & \textbf{72.51} & \textbf{78.58} & \textbf{75.95} & \textbf{91.48} & \textbf{81.84} \\ 
 & \cellcolor[gray]{0.9}$\Delta$Improve & \cellcolor[gray]{0.9}6.68\% & \cellcolor[gray]{0.9}16.05\% & \cellcolor[gray]{0.9}19.36\% & \cellcolor[gray]{0.9}13.20\% & \cellcolor[gray]{0.9}-0.38\% & \cellcolor[gray]{0.9}140.54\% & \cellcolor[gray]{0.9}12.95\% & \cellcolor[gray]{0.9}28.49\% & \cellcolor[gray]{0.9}10.18\% & \cellcolor[gray]{0.9}46.70\% & \cellcolor[gray]{0.9}22.96\% & \cellcolor[gray]{0.9}24.26\% \\ \midrule 
\multirow{4}{*}{Qwen2.5-7B-Instruct} 
 & zero-shot\textit{\ w/o chains text} & 23.61 & 42.54 & 14.73 & 27.39 & 11.27 & 44.92 & 19.81 & 24.81 & 31.71 & 39.45 & 15.82 & 29.17 \\
 & zero-shot & 53.08 & 45.32 & 11.41 & 38.59 & 22.22 & 48.23 & 1.21 & 24.34 & 40.81 & 48.32 & 1.75 & 30.78 \\
 & LoRA \textit{\ w/o chains text} & 62.82 & 58.59 & 71.97 & 64.03 & 31.28 & 52.11 & 12.41 & 32.36 & 55.33 & 68.65 & 85.89 & 69.52 \\
 & LoRA  & 74.60 & 65.64 & 75.62 & 71.90 & 22.39 & 56.61 & 66.79 & 46.95 & 66.83 & 70.95 & 84.09 & 73.72 \\ 
 & \textbf{GETER} & \textbf{76.41} & \textbf{74.61} & \textbf{84.49} & \textbf{78.12} & \textbf{63.77} & \textbf{70.06} & \textbf{88.42} & \textbf{73.27} & \textbf{78.23} & \textbf{72.95} & \textbf{89.90} & \textbf{80.23} \\ 
 & \cellcolor[gray]{0.9}$\Delta$Improve & \cellcolor[gray]{0.9}2.43\% & \cellcolor[gray]{0.9}13.66\% & \cellcolor[gray]{0.9}11.73\% & \cellcolor[gray]{0.9}8.65\% & \cellcolor[gray]{0.9}184.86\% & \cellcolor[gray]{0.9}23.77\% & \cellcolor[gray]{0.9}32.39\% & \cellcolor[gray]{0.9}56.04\% & \cellcolor[gray]{0.9}17.06\% & \cellcolor[gray]{0.9}2.82\% & \cellcolor[gray]{0.9}6.91\% & \cellcolor[gray]{0.9}8.83\% \\ \midrule 
\multirow{4}{*}{Mistral-7B-Instruct} 
 & zero-shot\textit{\ w/o chains text} & 3.65 & 39.44 & 46.44 & 27.81 & 5.52 & 40.64 & 23.50 & 22.39 & 7.56 & 33.21 & 46.10 & 28.37 \\
 & zero-shot & 22.04 & 27.64 & 40.76 & 29.26 & 0.99 & 24.93 & 43.23 & 21.55 & 17.73 & 29.80 & 49.69 & 31.96 \\
 & LoRA \textit{\ w/o chains text} & 58.04 & 65.44 & 80.03 & 66.79 & 19.45 & 58.16 & 71.52 & 47.80 & 70.81 & 39.12 & 75.80 & 61.95 \\
 & LoRA & 72.96 & 66.49 & 74.28 & 71.18 & 60.56 & 55.09 & 81.29 & 65.05 & 72.53 & 71.95 & 84.18 & 76.07 \\ 
 & \textbf{GETER} & \textbf{77.45} & \textbf{75.73} & \textbf{85.15} & \textbf{79.08} & \textbf{61.29} & \textbf{68.92} & \textbf{88.59} & \textbf{72.02} & \textbf{78.94} & \textbf{76.48} & \textbf{90.38} & \textbf{81.80}  \\
 & \cellcolor[gray]{0.9}$\Delta$Improve & \cellcolor[gray]{0.9}6.15\% & \cellcolor[gray]{0.9}13.89\% & \cellcolor[gray]{0.9}14.63\% & \cellcolor[gray]{0.9}11.10\% & \cellcolor[gray]{0.9}1.21\% & \cellcolor[gray]{0.9}25.11\% & \cellcolor[gray]{0.9}8.98\% & \cellcolor[gray]{0.9}10.71\% & \cellcolor[gray]{0.9}8.84\% & \cellcolor[gray]{0.9}6.30\% & \cellcolor[gray]{0.9}7.36\% & \cellcolor[gray]{0.9}7.54\% \\
\bottomrule
\end{tabular}
}
\end{center}
\caption{F1 scores (\%) of each model on the ICEWS14, GDELT, and ICEWS05-15 test instances. "Overall" represents the weighted average F1 score. \textit{w/o chains text} refers to the absence of reasoning chain input for LLMs. The best-performing results are highlighted in \textbf{bold}. $\Delta$Improve represents the relative improvements of \textbf{GETER} compared to \textbf{Tuned-only} methods. Additional datasets and detailed prediction results are provided in Appendix~\ref{sec:Full Experimental Results}.}
\label{tab: Main results classification metrics}
\end{table*}

\begin{table*}[htb]
\begin{center}
\resizebox{1.0\textwidth}{!}{
\begin{tabular}{c|c|cccc|cccc|cccc} 
\toprule
\multirow{2}{*}{Models} & \multirow{2}{*}{\diagbox{Types}{Datasets}} & \multicolumn{4}{c|}{ICEWS14} & \multicolumn{4}{c|}{GDELT} & \multicolumn{4}{c}{ICEWS05-15} \\ \cline{3-14}
 &  & BLEU-4 & rougeL & METEOR & BertScore (F1) & BLEU-4 & rougeL & METEOR & BertScore (F1) & BLEU-4 & rougeL & METEOR & BertScore (F1) \\ \midrule
\multirow{2}{*}{GPT-4o} 
 & zero-shot\textit{\ w/o chains text} & 10.78 & 23.82 & 31.14 & 68.16 & 5.95 & 21.30 & 26.84 & 64.73 & 10.74 & 23.63 & 30.94 & 68.00 \\
 & zero-shot & 22.94 & 41.04 & 37.24 & 79.25 & 9.16 & 27.61 & 32.32 & 70.91 & 22.64 & 40.83 & 36.27 & 79.16 \\ \midrule
\multirow{4}{*}{Llama3-8B-Instruct} 
 & zero-shot\textit{\ w/o chains text} & 4.35 & 16.32 & 16.71 & 61.35 & 2.38 & 13.41 & 17.03 & 56.98 & 2.27 & 12.88 & 10.53 & 58.28 \\
 & zero-shot & 9.70 & 30.19 & 26.60 & 70.25  & 5.61 & 27.10 & 25.73 & 67.42 & 10.08 & 31.13 & 27.44 & 70.02 \\
 & LoRA \textit{\ w/o chains text} & 27.73 & 39.71 & 45.94 & 80.16 & 18.12 & 37.05 & 35.92 & 77.51 & 27.59 & 39.63 & 45.80 & 80.17 \\
 & LoRA & 39.21 & 50.96 & \textbf{54.03} & 84.28 & 34.32 & 54.84 & 51.49 & \textbf{83.75} & 42.98 & 54.50 & 56.65 & 85.45  \\ 
 & \textbf{GETER} & \textbf{40.54} & \textbf{52.54} & 53.87 & \textbf{84.75} & \textbf{34.46} & \textbf{55.42} & \textbf{51.75} & 83.62 & \textbf{45.98} & \textbf{57.27} & \textbf{58.16} & \textbf{86.39} \\ 
 & \cellcolor[gray]{0.9}$\Delta$Improve & \cellcolor[gray]{0.9}3.39\% & \cellcolor[gray]{0.9}3.10\% & \cellcolor[gray]{0.9}-0.30\% & \cellcolor[gray]{0.9}0.56\% & \cellcolor[gray]{0.9}0.41\% & \cellcolor[gray]{0.9}1.06\% & \cellcolor[gray]{0.9}0.50\% & \cellcolor[gray]{0.9}-0.16\% & \cellcolor[gray]{0.9}6.98\% & \cellcolor[gray]{0.9}5.08\% & \cellcolor[gray]{0.9}2.67\% & \cellcolor[gray]{0.9}1.10\% \\\midrule   
 \multirow{4}{*}{Qwen2.5-7B-Instruct} 
 & zero-shot\textit{\ w/o chains text} & 7.43 & 19.73 & 30.82 & 66.03 & 3.76 & 17.90 & 28.25 & 63.15 & 7.81 & 19.87 & 30.27 & 65.94 \\
 & zero-shot & 11.18 & 28.49 & 27.98 & 72.28 & 7.55 & 26.90 & 25.97 & 70.00 & 10.53 & 28.53 & 26.32 & 72.04  \\
 & LoRA \textit{\ w/o chains text} & 28.17 & 40.22 & 45.20 & 80.12 & 17.15 & 36.89 & 34.52 & 75.71 & 28.60 & 40.52 & 45.76 & 80.39 \\
 & LoRA & 39.59 & \textbf{51.48} & 53.30 & 84.35 & 26.10 & 47.30 & 43.85 & 79.93 & 43.55 & 55.01 & 56.22 & 85.62 \\ 
 & \textbf{GETER} & \textbf{39.78} & 51.46 & \textbf{55.03} & \textbf{84.53} & \textbf{33.81} & \textbf{54.76} & \textbf{50.18} & \textbf{83.59} & \textbf{44.72} & \textbf{56.17} & \textbf{57.22} & \textbf{86.01} \\ 
 & \cellcolor[gray]{0.9}$\Delta$Improve & \cellcolor[gray]{0.9}0.48\% & \cellcolor[gray]{0.9}-0.04\% & \cellcolor[gray]{0.9}3.25\% & \cellcolor[gray]{0.9}0.21\% & \cellcolor[gray]{0.9}29.54\% & \cellcolor[gray]{0.9}15.76\% & \cellcolor[gray]{0.9}14.44\% & \cellcolor[gray]{0.9}4.58\% & \cellcolor[gray]{0.9}2.69\% & \cellcolor[gray]{0.9}2.11\% & \cellcolor[gray]{0.9}1.78\% & \cellcolor[gray]{0.9}0.46\% \\ \midrule   
\multirow{4}{*}{Mistral-7B-Instruct} 
 & zero-shot\textit{\ w/o chains text} & 7.17 & 19.40 & 24.27 & 65.46 & 4.89 & 18.20 & 25.78 & 63.60 & 7.24 & 19.29 & 23.26 & 65.10  \\
 & zero-shot & 9.19 & 28.36 & 25.70 & 71.63 & 7.46 & 27.96 & 25.99 & 70.43 & 7.95 & 27.40 & 23.60 & 70.73 \\
 & LoRA \textit{\ w/o chains text} & 28.01 & 39.84 & 45.70 & 80.34  & 18.22 & 38.08 & 35.74 & 76.76 & 28.26 & 40.13 & 45.96 & 80.45 \\
 & LoRA & 38.81 & 50.81 & 52.62 & 84.02 & 30.93 & 52.24 & 47.28 & 82.28 & 43.03 & 54.56 & 55.94 & 85.47 \\ 
 & \textbf{GETER} & \textbf{40.21} & \textbf{51.84} & \textbf{54.90} & \textbf{84.65} & \textbf{32.18} & \textbf{53.27} & \textbf{49.06} & \textbf{82.83} & \textbf{45.07} & \textbf{56.48} & \textbf{57.70} & \textbf{86.13} \\ 
 & \cellcolor[gray]{0.9}$\Delta$Improve & \cellcolor[gray]{0.9}3.61\% & \cellcolor[gray]{0.9}2.03\% & \cellcolor[gray]{0.9}4.33\% & \cellcolor[gray]{0.9}0.75\% & \cellcolor[gray]{0.9}4.04\% & \cellcolor[gray]{0.9}1.97\% & \cellcolor[gray]{0.9}3.77\% & \cellcolor[gray]{0.9}0.67\% & \cellcolor[gray]{0.9}4.74\% & \cellcolor[gray]{0.9}3.52\% & \cellcolor[gray]{0.9}3.14\% & \cellcolor[gray]{0.9}0.77\% \\ 
\bottomrule
\end{tabular}
}
\end{center}
\caption{The semantic similarity performance (\%) of each model on the ICEWS14, GDELT, and ICEWS05-15 test instances. \textit{w/o chains text} refers to the absence of reasoning chain input for LLMs. The best-performing results are highlighted in \textbf{bold}. Additional dataset explanation results are presented in Appendix~\ref{sec:Full Experimental Results}.}
\label{tab: Main results generate metrics}
\end{table*}

\subsection{Main results}
\label{sec: Main results}
In our experiments, we compare GETER with two model configurations: 1) \textit{Inference-only (zero-shot)}: Utilizing a frozen LLM to generate explanations directly without any additional training. 2) \textit{Tuned-only}: Fine-tuning the LLM using LoRA to enhance its performance on the task. Table~\ref{tab: Main results classification metrics} presents the prediction results, while Table~\ref{tab: Main results generate metrics} summarizes the explanation results. Overall, GETER demonstrates consistent and significant improvements across most metrics on both datasets, highlighting the effectiveness of the proposed approach. Further comparisons with graph-based methods are provided in Appendix~\ref{Comparison with Graph-based Methods}.

\textbf{Prediction Results.} Table~\ref{tab: Main results classification metrics} reports the prediction evaluation metrics for each LLM. The results show that both the \textit{Tuned-only} setting and GETER methods significantly outperform \textit{Inference-only} setting methods. This performance gap arises because fine-tuning allows models to better capture task-specific temporal patterns and improve logical consistency. Notably, GETER with Mistral demonstrates substantial improvements of $97.95\%$, $95.55\%$, and $101.58\%$ in overall F1 scores compared to the best-performing \textit{Inference-only} model GPT-4o. Furthermore, compared to \textit{Tuned-only} methods, GETER with Mistral achieves overall F1 score improvements of $11.10\%$, $10.71\%$, and $7.54\%$ across the three datasets. These results further underscore that GETER can effectively leverage the structural information of TKGs to enhance its explainable temporal reasoning capabilities.

\textbf{Explanation Results.} Table~\ref{tab: Main results generate metrics} presents the evaluation metrics for explanation generation. GETER demonstrates remarkable improvements across all key metrics. Specifically, compared to GPT-4o, GETER with Mistral achieves substantial enhancements in BLEU-4 scores across the three datasets, with gains of $75.28\%$, $251.31\%$, and $99.07\%$, respectively. These results highlight GETER's ability to leverage high-quality fine-tuning datasets to enhance explainable temporal reasoning capabilities.

\begin{table}[ht]
 \centering
 \setlength{\tabcolsep}{2mm}
 \resizebox{\linewidth}{!}{
 \begin{tabular}{cllll}\toprule
    \multirow{1}{*}{\textbf{No.}} & \multirow{1}{*}{\textbf{Model}} & \multicolumn{1}{c}{\textbf{ICEWS14}} & \multicolumn{1}{c}{\textbf{GDELT}} & \multicolumn{1}{c}{\textbf{ICEWS05-15}} \\
    \midrule
    1 & \textbf{GETER} & \textbf{79.08} & \textbf{72.02} & \textbf{81.80} \\
    \specialrule{0em}{1pt}{1pt}
    2 & GETER \textit{\ w/o} STA  & $71.18_{(\downarrow 7.90)}$ & $65.05_{(\downarrow 6.97)}$ & $76.07_{(\downarrow 5.73)}$ \\ 
    3 & GETER \textit{\ w/o} RCT & $72.05_{(\downarrow 7.03)}$ & $68.89_{(\downarrow 3.13)}$ & $77.82_{(\downarrow 3.98)}$ \\
    4 & GETER \textit{\ w/o} (STA \& RCT)  & $66.79_{(\downarrow 12.29)}$  & $47.80_{(\downarrow 24.22)}$  & $61.95_{(\downarrow 19.85)}$ \\
    \bottomrule
 \end{tabular}}
 \caption{Ablation study of GETER with Mistral on ICEWS14, GDELT, and ICEWS05-15 datasets using overall F1 scores (\%). STA denotes structure-text adapter, while RCT denotes reasoning chains text.}
 \label{tab: Ablation Study}
\end{table}

\subsection{Ablation Study}

In this subsection, we conduct an ablation study to investigate the individual contributions of different components in GETER. The results for various variants are presented in Table~\ref{tab: Ablation Study}, indicating that all modules are essential, as removing any of them leads to a decline in performance. Notably, to validate the usefulness of the structural information provided by GETER, we directly removed the structure-text adapter from the model (Line 2). This ablation results in overall F1 score reductions of $11.10\%$, $10.71\%$, and $7.53\%$ across the three datasets, respectively. These results demonstrate that the soft graph token with lightweight adapter can effective capture the structural characteristics for the query. Additionally, as shown in Line 3 of Table~\ref{tab: Ablation Study}, removing the reasoning chains text leads to a significant performance decline, with F1 scores dropping by $9.76\%$, $10.71\%$, and $5.11\%$ across the three datasets, respectively. This result highlights the importance of reasoning chains text, as they provide sequenced evidence that enriches the contextual background. Furthermore, we observe that GETER scheme significantly outperforms the base model that directly adopts instruction tuning (Line 4). This demonstrates the effectiveness of GETER, which combine structural and contextual semantic information to activate and harness the LLM's capability for explainable temporal reasoning.

\subsection{Discussion}
\label{Discussion}

In this subsection, we conduct further analysis of the impact of different temporal encoders, the influence of MLP depth, and the effect of various reasoning chain serialization formats on the model's performance. All experiments are conducted using Mistral for its superior performance.  Additionally, we present a complexity analysis in Appendix~\ref{Complexity Analysis of GETER} and a case study in Appendix~\ref{sec:Case Study} to further highlight the advantages of our proposed method.

\begin{figure}[ht]
    \centering
    \subfloat[ICEWS14]{\includegraphics[width=1.5in]{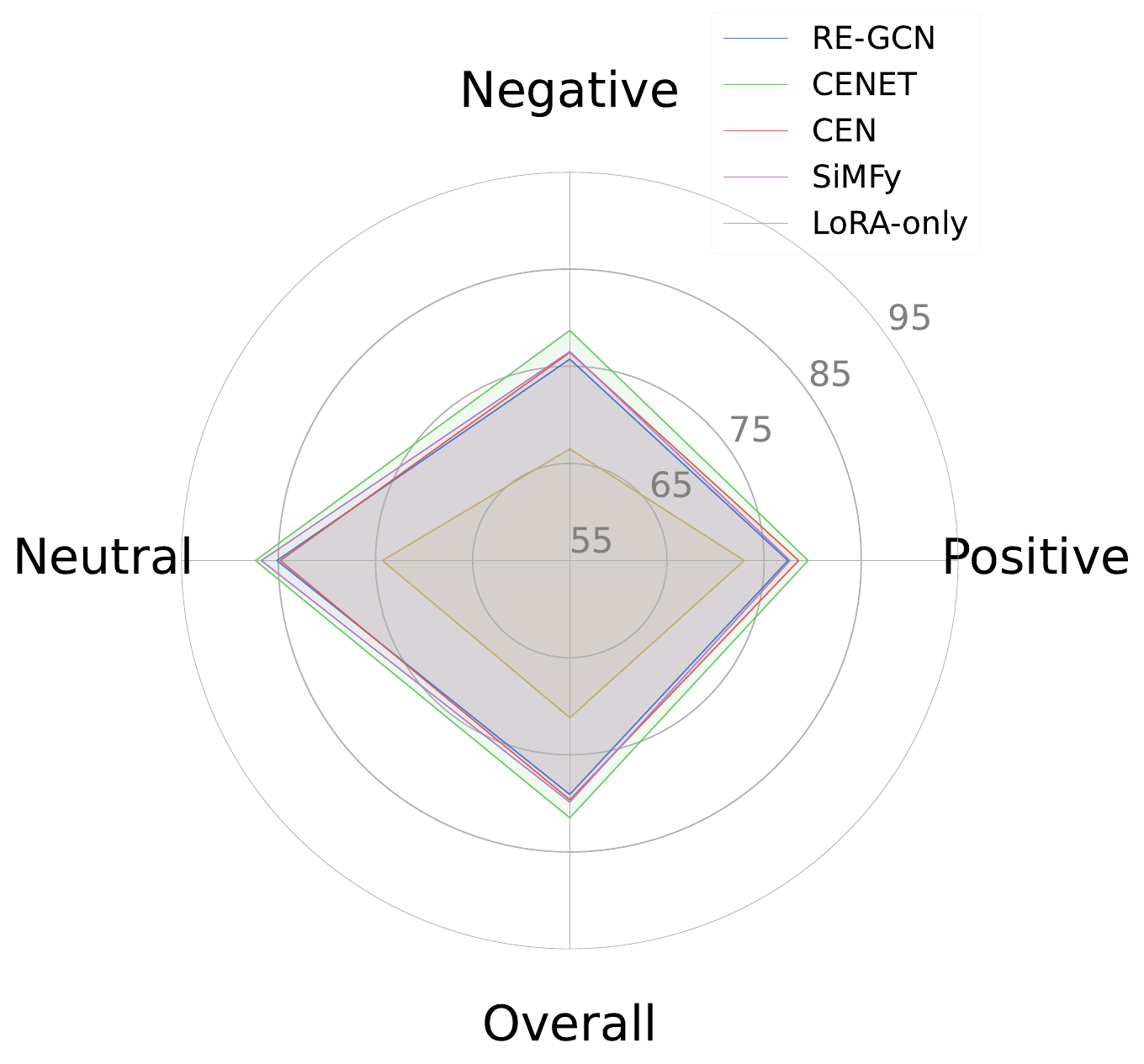} \label{ICEWS14}} 
    \subfloat[GDELT]{\includegraphics[width=1.5in]{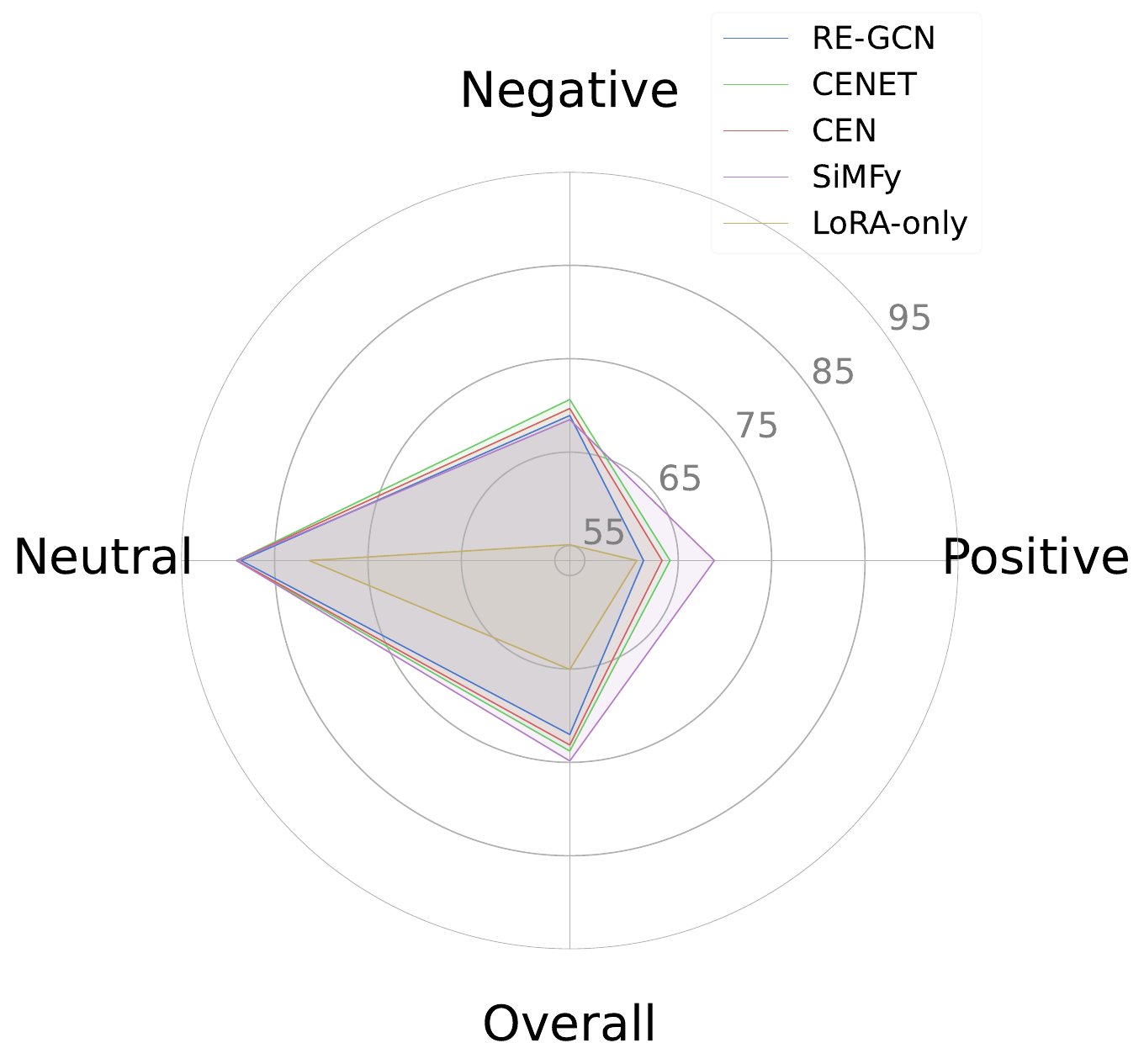} \label{GDELT}} 
    \caption{Comparison of GETER with different temporal encoders on the ICEWS14 and GDELT datasets in terms of overall F1 scores (\%).}
    \label{fig:temporal_encoder}
\end{figure}

\noindent \textbf{Q1: What is the impact of different temporal encoders on GETER’s performance?} To evaluate the impact of different temporal encoders, we also integrate CEN, CENET, and SiMFy into the our framework, as described in~\ref{Experiments Setup}. The performance comparison is illustrated in Figure~\ref{fig:temporal_encoder}. The results demonstrate that GETER achieves consistently high performance across two datasets when paired with any of the three temporal encoders, significantly outperforming methods that rely solely on LoRA. These findings demonstrate that GETER is robust to variations in temporal encoders. Details about temporal encoders refer to Appendix~\ref{sec:Baselines}.

\begin{figure}[!t]
    \centering
    \includegraphics[width=0.95\linewidth]{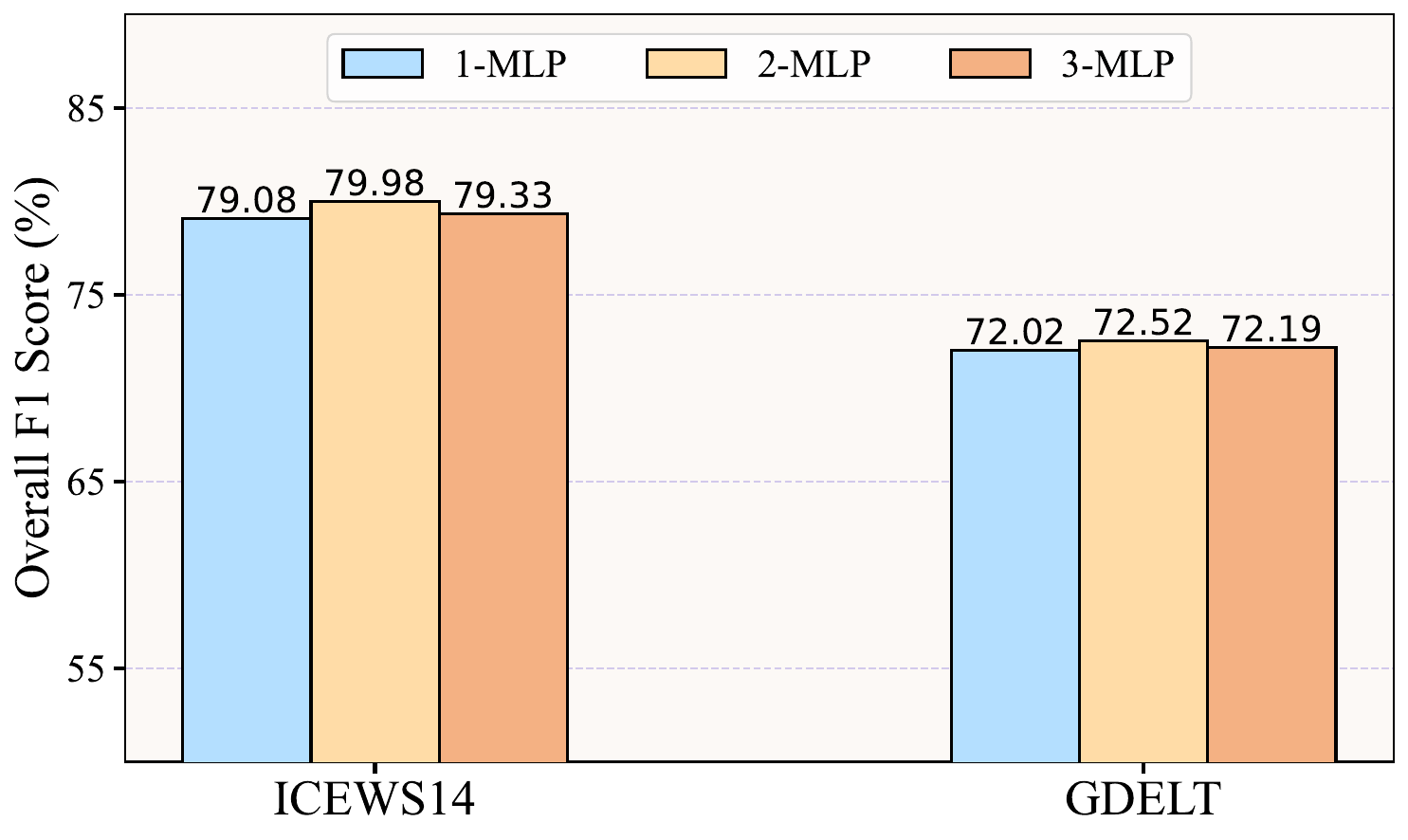}
    \caption{MLP depth comparison on ICEWS14 and GDELT datasets in terms of overall F1 scores (\%).}
    \label{fig:Discuss on depth of MLPs}
\end{figure}

\noindent \textbf{Q2: How does the depth of the MLP affect GETER’s performance?} GETER uses a one-layer MLP to map the graph structure feature into the text embedding space. To investigate whether deeper neural structures improves performance, we conduct experiments to replace the one-layer MLP with deeper variants. The results on the ICEWS14 and GDELT datasets are presented in Figure~\ref{fig:Discuss on depth of MLPs}. We can observe that increasing model complexity has minimal impact on performance. This is likely because deeper structures fail to capture evolving structural information more effectively.

\begin{table}[!t]
\centering
\setlength{\tabcolsep}{1.4mm}
\resizebox{\linewidth}{!}{
\begin{tabular}{lcccc}\toprule
    \textbf{\textit{Model}} & \textbf{Positive} & \textbf{Negative} & \textbf{Neutral} & \textbf{Overall}\\
    \midrule
    \textit{\textbf{GETER} (paths order)}   & 77.45 & 75.73 & 85.15 & 79.08 \\ \hline
    \textit{descending order}   &  \textbf{80.53} & 76.00 & \textbf{86.34} & \textbf{80.68} \\
    \textit{ascending order}  & 77.72 & \textbf{77.52} & 86.04 & 80.03 \\
    \textit{random order}  & 75.02 & 76.31 & 82.45 & 77.57 \\
    \bottomrule
\end{tabular}}
\caption{Performance (F1 (\%)) of GETER with different reasoning chain formats on the ICEWS14 dataset.}
\label{tab:different reasoning chain text formats}
\end{table}

\noindent \textbf{Q3: What is the effect of different reasoning chain text formats on GETER’s performance?} We further investigate how GETER utilizes reasoning chain text, which provides contextualized background information for queries. Specifically, we evaluate three different serialization formats based on the timestamp of quadruples: \textit{ascending}, \textit{descending}, and \textit{random}. As shown in Table~\ref{tab:different reasoning chain text formats}, the model achieves the best performance with the descending order format. Surprisingly, even with random serialization, GETER still maintains competitive performance. This is attributed to the structured adapter in GETER, which effectively couple structure and text information in a contextualized manner. These findings further highlight the robustness and adaptability of our proposed GETER.

\section{Conclusion}

We introduce a comprehensive benchmark covering a wide range of temporal granularities for systematically evaluating LLMs' explainable temporal reasoning. To address the challenge of LLMs struggling to deliver convincing explanations, we propose a novel structure-aware generative framework \textbf{GETER}, which effectively bridges the gap between graph structures and text by through a lightweight structure-text adapter. Extensive experiments validate the effectiveness and robustness of our proposed GETER. 
 
\section*{Limitations}
GETER can effectively activate and harness the explainable reasoning ability of LLMs by incorporate the graph structural information into the LLMs. However, the extremely large number of parameters in LLMs makes fine-tuning them resource-intensive. At the same time, LLMs are notoriously slow at decoding during inference. In our experiment, we use DeepSpeed~\cite{DBLP:conf/sc/RajbhandariRRH20} to accelerate training and inference. Additionally, some reasoning chains may introduce noisy text, which could negatively affect explainable temporal reasoning performance.

\section*{Ethics Statement}

In developing this explainable temporal reasoning benchmark, all data used in this study are publicly available and do not pose any privacy concerns. Additionally,  we have carefully considered ethical issues and limitations commonly associated with large language models. Nonetheless, we acknowledge that, despite our best efforts, the benchmark may still contain gaps or unintended biases. To mitigate this, the source data has been meticulously curated to ensure diversity and minimize potential biases. Through rigorous design and testing processes, we strive to uphold ethical AI principles while advancing research in temporal reasoning.

\section*{Acknowledgments}

We would like to thank all the anonymous reviewers and area chairs for their comments. This research is supported by National Natural Science Foundation of China (U23A20316) and founded by Joint\&Laboratory on Credit Technology.
\bibliography{acl_latex}

\begin{thebibliography}{47}
\providecommand{\natexlab}[1]{#1}

\bibitem[{Banerjee and Lavie(2005)}]{METEOR}
Satanjeev Banerjee and Alon Lavie. 2005.
\newblock {METEOR:} an automatic metric for {MT} evaluation with improved correlation with human judgments.
\newblock In \emph{IEEvaluation@ACL}, pages 65--72. Association for Computational Linguistics.

\bibitem[{Bogina et~al.(2023)Bogina, Kuflik, Jannach, Bielikov{\'{a}}, Kompan, and Trattner}]{DBLP:journals/umuai/BoginaKJBKT23}
Veronika Bogina, Tsvi Kuflik, Dietmar Jannach, M{\'{a}}ria Bielikov{\'{a}}, Michal Kompan, and Christoph Trattner. 2023.
\newblock \href {https://doi.org/10.1007/S11257-022-09335-W} {Considering temporal aspects in recommender systems: a survey}.
\newblock \emph{User Model. User Adapt. Interact.}, 33(1):81--119.

\bibitem[{Chu et~al.(2024)Chu, Chen, Chen, Yu, Wang, Liu, and Qin}]{DBLP:conf/acl/ChuCCY00024}
Zheng Chu, Jingchang Chen, Qianglong Chen, Weijiang Yu, Haotian Wang, Ming Liu, and Bing Qin. 2024.
\newblock Timebench: {A} comprehensive evaluation of temporal reasoning abilities in large language models.
\newblock In \emph{{ACL} {(1)}}, pages 1204--1228. Association for Computational Linguistics.

\bibitem[{Dubey et~al.(2024)Dubey, Jauhri, Pandey, Kadian, Al{-}Dahle, Letman, Mathur, Schelten, Yang, Fan, Goyal, Hartshorn, Yang, Mitra, Sravankumar, Korenev, Hinsvark, Rao, Zhang, Rodriguez, Gregerson, Spataru, Rozi{\`{e}}re, Biron, Tang, Chern, Caucheteux, Nayak, Bi, Marra, McConnell, Keller, Touret, Wu, Wong, Ferrer, Nikolaidis, Allonsius, Song, Pintz, Livshits, Esiobu, Choudhary, Mahajan, Garcia{-}Olano, Perino, Hupkes, Lakomkin, AlBadawy, Lobanova, Dinan, Smith, Radenovic, Zhang, Synnaeve, Lee, Anderson, Nail, Mialon, Pang, Cucurell, Nguyen, Korevaar, Xu, Touvron, Zarov, Ibarra, Kloumann, Misra, Evtimov, Copet, Lee, Geffert, Vranes, Park, Mahadeokar, Shah, van~der Linde, Billock, Hong, Lee, Fu, Chi, Huang, Liu, Wang, Yu, Bitton, Spisak, Park, Rocca, Johnstun, Saxe, Jia, Alwala, Upasani, Plawiak, Li, Heafield, Stone, and et~al.}]{Llama3-8b}
Abhimanyu Dubey, Abhinav Jauhri, Abhinav Pandey, Abhishek Kadian, Ahmad Al{-}Dahle, Aiesha Letman, Akhil Mathur, Alan Schelten, Amy Yang, Angela Fan, Anirudh Goyal, Anthony Hartshorn, Aobo Yang, Archi Mitra, Archie Sravankumar, Artem Korenev, Arthur Hinsvark, Arun Rao, Aston Zhang, Aur{\'{e}}lien Rodriguez, Austen Gregerson, Ava Spataru, Baptiste Rozi{\`{e}}re, Bethany Biron, Binh Tang, Bobbie Chern, Charlotte Caucheteux, Chaya Nayak, Chloe Bi, Chris Marra, Chris McConnell, Christian Keller, Christophe Touret, Chunyang Wu, Corinne Wong, Cristian~Canton Ferrer, Cyrus Nikolaidis, Damien Allonsius, Daniel Song, Danielle Pintz, Danny Livshits, David Esiobu, Dhruv Choudhary, Dhruv Mahajan, Diego Garcia{-}Olano, Diego Perino, Dieuwke Hupkes, Egor Lakomkin, Ehab AlBadawy, Elina Lobanova, Emily Dinan, Eric~Michael Smith, Filip Radenovic, Frank Zhang, Gabriel Synnaeve, Gabrielle Lee, Georgia~Lewis Anderson, Graeme Nail, Gr{\'{e}}goire Mialon, Guan Pang, Guillem Cucurell, Hailey Nguyen, Hannah Korevaar, Hu~Xu, Hugo
  Touvron, Iliyan Zarov, Imanol~Arrieta Ibarra, Isabel~M. Kloumann, Ishan Misra, Ivan Evtimov, Jade Copet, Jaewon Lee, Jan Geffert, Jana Vranes, Jason Park, Jay Mahadeokar, Jeet Shah, Jelmer van~der Linde, Jennifer Billock, Jenny Hong, Jenya Lee, Jeremy Fu, Jianfeng Chi, Jianyu Huang, Jiawen Liu, Jie Wang, Jiecao Yu, Joanna Bitton, Joe Spisak, Jongsoo Park, Joseph Rocca, Joshua Johnstun, Joshua Saxe, Junteng Jia, Kalyan~Vasuden Alwala, Kartikeya Upasani, Kate Plawiak, Ke~Li, Kenneth Heafield, Kevin Stone, and et~al. 2024.
\newblock The llama 3 herd of models.
\newblock \emph{CoRR}, abs/2407.21783.

\bibitem[{Garc{\'{\i}}a{-}Dur{\'{a}}n et~al.(2018)Garc{\'{\i}}a{-}Dur{\'{a}}n, Dumancic, and Niepert}]{DBLP:conf/emnlp/Garcia-DuranDN18}
Alberto Garc{\'{\i}}a{-}Dur{\'{a}}n, Sebastijan Dumancic, and Mathias Niepert. 2018.
\newblock Learning sequence encoders for temporal knowledge graph completion.
\newblock In \emph{Proc. of EMNLP}, pages 4816--4821.

\bibitem[{Han et~al.(2021)Han, Chen, Ma, and Tresp}]{xERTE}
Zhen Han, Peng Chen, Yunpu Ma, and Volker Tresp. 2021.
\newblock Explainable subgraph reasoning for forecasting on temporal knowledge graphs.
\newblock In \emph{{ICLR}}. OpenReview.net.

\bibitem[{He et~al.(2023)He, Gao, and Chen}]{NLI}
Pengcheng He, Jianfeng Gao, and Weizhu Chen. 2023.
\newblock \href {https://openreview.net/forum?id=sE7-XhLxHA} {Debertav3: Improving deberta using electra-style pre-training with gradient-disentangled embedding sharing}.
\newblock In \emph{The Eleventh International Conference on Learning Representations, {ICLR} 2023, Kigali, Rwanda, May 1-5, 2023}. OpenReview.net.

\bibitem[{He et~al.(2024)He, Tian, Sun, Chawla, Laurent, LeCun, Bresson, and Hooi}]{G-Retriever}
Xiaoxin He, Yijun Tian, Yifei Sun, Nitesh~V. Chawla, Thomas Laurent, Yann LeCun, Xavier Bresson, and Bryan Hooi. 2024.
\newblock \href {https://doi.org/10.48550/ARXIV.2402.07630} {G-retriever: Retrieval-augmented generation for textual graph understanding and question answering}.
\newblock \emph{CoRR}, abs/2402.07630.

\bibitem[{Hu et~al.(2022)Hu, Shen, Wallis, Allen{-}Zhu, Li, Wang, Wang, and Chen}]{LORA}
Edward~J. Hu, Yelong Shen, Phillip Wallis, Zeyuan Allen{-}Zhu, Yuanzhi Li, Shean Wang, Lu~Wang, and Weizhu Chen. 2022.
\newblock \href {https://openreview.net/forum?id=nZeVKeeFYf9} {Lora: Low-rank adaptation of large language models}.
\newblock In \emph{The Tenth International Conference on Learning Representations, {ICLR} 2022, Virtual Event, April 25-29, 2022}. OpenReview.net.

\bibitem[{Huang and Chang(2023)}]{DBLP:conf/acl/0009C23}
Jie Huang and Kevin~Chen{-}Chuan Chang. 2023.
\newblock Towards reasoning in large language models: {A} survey.
\newblock In \emph{{ACL} (Findings)}, pages 1049--1065. Association for Computational Linguistics.

\bibitem[{Jiang et~al.(2024)Jiang, Sablayrolles, Roux, Mensch, Savary, Bamford, Chaplot, de~Las~Casas, Hanna, Bressand, Lengyel, Bour, Lample, Lavaud, Saulnier, Lachaux, Stock, Subramanian, Yang, Antoniak, Scao, Gervet, Lavril, Wang, Lacroix, and Sayed}]{mistral}
Albert~Q. Jiang, Alexandre Sablayrolles, Antoine Roux, Arthur Mensch, Blanche Savary, Chris Bamford, Devendra~Singh Chaplot, Diego de~Las~Casas, Emma~Bou Hanna, Florian Bressand, Gianna Lengyel, Guillaume Bour, Guillaume Lample, L{\'{e}}lio~Renard Lavaud, Lucile Saulnier, Marie{-}Anne Lachaux, Pierre Stock, Sandeep Subramanian, Sophia Yang, Szymon Antoniak, Teven~Le Scao, Th{\'{e}}ophile Gervet, Thibaut Lavril, Thomas Wang, Timoth{\'{e}}e Lacroix, and William~El Sayed. 2024.
\newblock Mixtral of experts.
\newblock \emph{CoRR}, abs/2401.04088.

\bibitem[{Jiang et~al.(2023)Jiang, Zhou, Zhao, Li, and Wen}]{DBLP:conf/emnlp/JiangZZLW23}
Jinhao Jiang, Kun Zhou, Wayne~Xin Zhao, Yaliang Li, and Ji{-}Rong Wen. 2023.
\newblock \href {https://doi.org/10.18653/V1/2023.EMNLP-MAIN.228} {Reasoninglm: Enabling structural subgraph reasoning in pre-trained language models for question answering over knowledge graph}.
\newblock In \emph{Proceedings of the 2023 Conference on Empirical Methods in Natural Language Processing, {EMNLP} 2023, Singapore, December 6-10, 2023}, pages 3721--3735. Association for Computational Linguistics.

\bibitem[{Jin et~al.(2020)Jin, Qu, Jin, and Ren}]{RE-NET}
Woojeong Jin, Meng Qu, Xisen Jin, and Xiang Ren. 2020.
\newblock Recurrent event network: Autoregressive structure inferenceover temporal knowledge graphs.
\newblock In \emph{Proc. of EMNLP}, pages 6669--6683.

\bibitem[{Leblay and Chekol(2018)}]{DBLP:conf/www/LeblayC18}
Julien Leblay and Melisachew~Wudage Chekol. 2018.
\newblock \href {https://doi.org/10.1145/3184558.3191639} {Deriving validity time in knowledge graph}.
\newblock In \emph{Companion of the The Web Conference 2018 on The Web Conference 2018, {WWW} 2018, Lyon , France, April 23-27, 2018}, pages 1771--1776. {ACM}.

\bibitem[{Lee et~al.(2023)Lee, Ahrabian, Jin, Morstatter, and Pujara}]{ICL}
Dong{-}Ho Lee, Kian Ahrabian, Woojeong Jin, Fred Morstatter, and Jay Pujara. 2023.
\newblock Temporal knowledge graph forecasting without knowledge using in-context learning.
\newblock In \emph{{EMNLP}}, pages 544--557. Association for Computational Linguistics.

\bibitem[{Li et~al.(2022)Li, Guan, Jin, Peng, Lyu, Zhu, Bai, Li, Guo, and Cheng}]{CEN}
Zixuan Li, Saiping Guan, Xiaolong Jin, Weihua Peng, Yajuan Lyu, Yong Zhu, Long Bai, Wei Li, Jiafeng Guo, and Xueqi Cheng. 2022.
\newblock Complex evolutional pattern learning for temporal knowledge graph reasoning.
\newblock In \emph{{ACL} {(2)}}, pages 290--296. Association for Computational Linguistics.

\bibitem[{Li et~al.(2021)Li, Jin, Li, Guan, Guo, Shen, Wang, and Cheng}]{RE-GCN}
Zixuan Li, Xiaolong Jin, Wei Li, Saiping Guan, Jiafeng Guo, Huawei Shen, Yuanzhuo Wang, and Xueqi Cheng. 2021.
\newblock \href {https://doi.org/10.1145/3404835.3462963} {Temporal knowledge graph reasoning based on evolutional representation learning}.
\newblock In \emph{{SIGIR} '21: The 44th International {ACM} {SIGIR} Conference on Research and Development in Information Retrieval, Virtual Event, Canada, July 11-15, 2021}, pages 408--417. {ACM}.

\bibitem[{Liao et~al.(2024)Liao, Jia, Li, Ma, and Tresp}]{GenTKG}
Ruotong Liao, Xu~Jia, Yangzhe Li, Yunpu Ma, and Volker Tresp. 2024.
\newblock Gentkg: Generative forecasting on temporal knowledge graph with large language models.
\newblock In \emph{{NAACL-HLT} (Findings)}, pages 4303--4317. Association for Computational Linguistics.

\bibitem[{Lin(2004)}]{ROUGE}
Chin-Yew Lin. 2004.
\newblock \href {https://aclanthology.org/W04-1013/} {{ROUGE}: A package for automatic evaluation of summaries}.
\newblock In \emph{Text Summarization Branches Out}, pages 74--81, Barcelona, Spain. Association for Computational Linguistics.

\bibitem[{Lin et~al.(2023)Lin, Liu, Mao, Xu, and Cambria}]{TECHS}
Qika Lin, Jun Liu, Rui Mao, Fangzhi Xu, and Erik Cambria. 2023.
\newblock {TECHS:} temporal logical graph networks for explainable extrapolation reasoning.
\newblock In \emph{{ACL} {(1)}}, pages 1281--1293. Association for Computational Linguistics.

\bibitem[{Liu et~al.(2025{\natexlab{a}})Liu, Zhang, Lin, Yang, and Peng}]{Filter-then-Generate}
Ben Liu, Jihai Zhang, Fangquan Lin, Cheng Yang, and Min Peng. 2025{\natexlab{a}}.
\newblock \href {https://aclanthology.org/2025.coling-main.740/} {Filter-then-generate: Large language models with structure-text adapter for knowledge graph completion}.
\newblock In \emph{Proceedings of the 31st International Conference on Computational Linguistics, {COLING} 2025, Abu Dhabi, UAE, January 19-24, 2025}, pages 11181--11195. Association for Computational Linguistics.

\bibitem[{Liu et~al.(2025{\natexlab{b}})Liu, Zhang, Lin, Yang, Peng, and Yin}]{DBLP:conf/www/LiuZLYPY25}
Ben Liu, Jihai Zhang, Fangquan Lin, Cheng Yang, Min Peng, and Wotao Yin. 2025{\natexlab{b}}.
\newblock Symagent: {A} neural-symbolic self-learning agent framework for complex reasoning over knowledge graphs.
\newblock In \emph{{WWW}}, pages 98--108. {ACM}.

\bibitem[{Liu et~al.(2022{\natexlab{a}})Liu, Tam, Muqeeth, Mohta, Huang, Bansal, and Raffel}]{DBLP:conf/nips/LiuTMMHBR22}
Haokun Liu, Derek Tam, Mohammed Muqeeth, Jay Mohta, Tenghao Huang, Mohit Bansal, and Colin Raffel. 2022{\natexlab{a}}.
\newblock \href {http://papers.nips.cc/paper\_files/paper/2022/hash/0cde695b83bd186c1fd456302888454c-Abstract-Conference.html} {Few-shot parameter-efficient fine-tuning is better and cheaper than in-context learning}.
\newblock In \emph{Advances in Neural Information Processing Systems 35: Annual Conference on Neural Information Processing Systems 2022, NeurIPS 2022, New Orleans, LA, USA, November 28 - December 9, 2022}.

\bibitem[{Liu et~al.(2022{\natexlab{b}})Liu, Ma, Hildebrandt, Joblin, and Tresp}]{Tlogic}
Yushan Liu, Yunpu Ma, Marcel Hildebrandt, Mitchell Joblin, and Volker Tresp. 2022{\natexlab{b}}.
\newblock Tlogic: Temporal logical rules for explainable link forecasting on temporal knowledge graphs.
\newblock In \emph{{AAAI}}, pages 4120--4127. {AAAI} Press.

\bibitem[{Liu et~al.(2023)Liu, Tan, Li, Wan, Jin, and Shi}]{SiMFy}
Zhengtao Liu, Lei Tan, Mengfan Li, Yao Wan, Hai Jin, and Xuanhua Shi. 2023.
\newblock Simfy: {A} simple yet effective approach for temporal knowledge graph reasoning.
\newblock In \emph{{EMNLP} (Findings)}, pages 3825--3836. Association for Computational Linguistics.

\bibitem[{Luo et~al.(2024)Luo, Gu, Li, Li, Lin, Li, and Yang}]{COH2}
Ruilin Luo, Tianle Gu, Haoling Li, Junzhe Li, Zicheng Lin, Jiayi Li, and Yujiu Yang. 2024.
\newblock Chain of history: Learning and forecasting with llms for temporal knowledge graph completion.
\newblock \emph{CoRR}, abs/2401.06072.

\bibitem[{Ma et~al.(2024)Ma, Ren, and Huang}]{DBLP:conf/emnlp/MaR024}
Qiyao Ma, Xubin Ren, and Chao Huang. 2024.
\newblock Xrec: Large language models for explainable recommendation.
\newblock In \emph{{EMNLP} (Findings)}, pages 391--402. Association for Computational Linguistics.

\bibitem[{Mei et~al.(2022)Mei, Yang, Cai, and Jiang}]{DBLP:conf/emnlp/MeiYCJ22}
Xin Mei, Libin Yang, Xiaoyan Cai, and Zuowei Jiang. 2022.
\newblock An adaptive logical rule embedding model for inductive reasoning over temporal knowledge graphs.
\newblock In \emph{{EMNLP}}, pages 7304--7316. Association for Computational Linguistics.

\bibitem[{OpenAI(2023)}]{GPT-4o}
OpenAI. 2023.
\newblock {GPT-4} technical report.
\newblock \emph{CoRR}, abs/2303.08774.

\bibitem[{Papineni et~al.(2002)Papineni, Roukos, Ward, and Zhu}]{BLEU}
Kishore Papineni, Salim Roukos, Todd Ward, and Wei{-}Jing Zhu. 2002.
\newblock Bleu: a method for automatic evaluation of machine translation.
\newblock In \emph{{ACL}}, pages 311--318. {ACL}.

\bibitem[{Peng et~al.(2025)Peng, Chen, Suo, and Li}]{DBLP:journals/corr/abs-2503-00845}
Miao Peng, Nuo Chen, Zongrui Suo, and Jia Li. 2025.
\newblock Rewarding graph reasoning process makes llms more generalized reasoners.
\newblock \emph{CoRR}, abs/2503.00845.

\bibitem[{Rajbhandari et~al.(2020)Rajbhandari, Rasley, Ruwase, and He}]{DBLP:conf/sc/RajbhandariRRH20}
Samyam Rajbhandari, Jeff Rasley, Olatunji Ruwase, and Yuxiong He. 2020.
\newblock \href {https://doi.org/10.1109/SC41405.2020.00024} {Zero: memory optimizations toward training trillion parameter models}.
\newblock In \emph{Proceedings of the International Conference for High Performance Computing, Networking, Storage and Analysis, {SC} 2020, Virtual Event / Atlanta, Georgia, USA, November 9-19, 2020}, page~20. {IEEE/ACM}.

\bibitem[{Sun et~al.(2021)Sun, Zhong, Ma, Han, and He}]{TITer}
Haohai Sun, Jialun Zhong, Yunpu Ma, Zhen Han, and Kun He. 2021.
\newblock Timetraveler: Reinforcement learning for temporal knowledge graph forecasting.
\newblock In \emph{{EMNLP} {(1)}}, pages 8306--8319. Association for Computational Linguistics.

\bibitem[{Tan et~al.(2023)Tan, Ng, and Bing}]{relatedwork_2}
Qingyu Tan, Hwee~Tou Ng, and Lidong Bing. 2023.
\newblock Towards benchmarking and improving the temporal reasoning capability of large language models.
\newblock In \emph{{ACL} {(1)}}, pages 14820--14835. Association for Computational Linguistics.

\bibitem[{Tan et~al.(2024)Tan, Ng, and Bing}]{relatedwork_7}
Qingyu Tan, Hwee~Tou Ng, and Lidong Bing. 2024.
\newblock Towards robust temporal reasoning of large language models via a multi-hop {QA} dataset and pseudo-instruction tuning.
\newblock In \emph{{ACL} (Findings)}, pages 6272--6286. Association for Computational Linguistics.

\bibitem[{Wang et~al.(2023)Wang, Wei, Schuurmans, Le, Chi, Narang, Chowdhery, and Zhou}]{DBLP:conf/iclr/0002WSLCNCZ23}
Xuezhi Wang, Jason Wei, Dale Schuurmans, Quoc~V. Le, Ed~H. Chi, Sharan Narang, Aakanksha Chowdhery, and Denny Zhou. 2023.
\newblock Self-consistency improves chain of thought reasoning in language models.
\newblock In \emph{{ICLR}}. OpenReview.net.

\bibitem[{Wang and Zhao(2024)}]{DBLP:conf/acl/Wang024}
Yuqing Wang and Yun Zhao. 2024.
\newblock {TRAM:} benchmarking temporal reasoning for large language models.
\newblock In \emph{{ACL} (Findings)}, pages 6389--6415. Association for Computational Linguistics.

\bibitem[{Wei et~al.(2022)Wei, Tay, Bommasani, Raffel, Zoph, Borgeaud, Yogatama, Bosma, Zhou, Metzler, Chi, Hashimoto, Vinyals, Liang, Dean, and Fedus}]{DBLP:journals/tmlr/WeiTBRZBYBZMCHVLDF22}
Jason Wei, Yi~Tay, Rishi Bommasani, Colin Raffel, Barret Zoph, Sebastian Borgeaud, Dani Yogatama, Maarten Bosma, Denny Zhou, Donald Metzler, Ed~H. Chi, Tatsunori Hashimoto, Oriol Vinyals, Percy Liang, Jeff Dean, and William Fedus. 2022.
\newblock Emergent abilities of large language models.
\newblock \emph{Trans. Mach. Learn. Res.}, 2022.

\bibitem[{Wei et~al.(2023)Wei, Su, Ma, Yu, Lei, Zhang, Zhao, and Liu}]{relatedwork_4}
Yifan Wei, Yisong Su, Huanhuan Ma, Xiaoyan Yu, Fangyu Lei, Yuanzhe Zhang, Jun Zhao, and Kang Liu. 2023.
\newblock Menatqa: {A} new dataset for testing the temporal comprehension and reasoning abilities of large language models.
\newblock In \emph{{EMNLP} (Findings)}, pages 1434--1447. Association for Computational Linguistics.

\bibitem[{Wu et~al.(2025)Wu, Huang, Jiang, Xie, Huang, and Zhao}]{DBLP:journals/corr/abs-2501-00888}
Weiqi Wu, Shen Huang, Yong Jiang, Pengjun Xie, Fei Huang, and Hai Zhao. 2025.
\newblock \href {https://doi.org/10.48550/ARXIV.2501.00888} {Unfolding the headline: Iterative self-questioning for news retrieval and timeline summarization}.
\newblock \emph{CoRR}, abs/2501.00888.

\bibitem[{Xia et~al.(2024)Xia, Wang, Liu, Wang, Wu, and Zhang}]{COH}
Yuwei Xia, Ding Wang, Qiang Liu, Liang Wang, Shu Wu, and Xiaoyu Zhang. 2024.
\newblock Chain-of-history reasoning for temporal knowledge graph forecasting.
\newblock In \emph{{ACL} (Findings)}, pages 16144--16159. Association for Computational Linguistics.

\bibitem[{Xiong et~al.(2024)Xiong, Payani, Kompella, and Fekri}]{relatedwork_3}
Siheng Xiong, Ali Payani, Ramana Kompella, and Faramarz Fekri. 2024.
\newblock Large language models can learn temporal reasoning.
\newblock In \emph{{ACL} {(1)}}, pages 10452--10470. Association for Computational Linguistics.

\bibitem[{Xu et~al.(2023)Xu, Ou, Xu, and Fu}]{CENET}
Yi~Xu, Junjie Ou, Hui Xu, and Luoyi Fu. 2023.
\newblock Temporal knowledge graph reasoning with historical contrastive learning.
\newblock In \emph{{AAAI}}, pages 4765--4773. {AAAI} Press.

\bibitem[{Yang et~al.(2024{\natexlab{a}})Yang, Yang, Hui, Zheng, Yu, Zhou, Li, Li, Liu, Huang, Dong, Wei, Lin, Tang, Wang, Yang, Tu, Zhang, Ma, Xu, Zhou, Bai, He, Lin, Dang, Lu, Chen, Yang, Li, Xue, Ni, Zhang, Wang, Peng, Men, Gao, Lin, Wang, Bai, Tan, Zhu, Li, Liu, Ge, Deng, Zhou, Ren, Zhang, Wei, Ren, Fan, Yao, Zhang, Wan, Chu, Liu, Cui, Zhang, and Fan}]{qwen2}
An~Yang, Baosong Yang, Binyuan Hui, Bo~Zheng, Bowen Yu, Chang Zhou, Chengpeng Li, Chengyuan Li, Dayiheng Liu, Fei Huang, Guanting Dong, Haoran Wei, Huan Lin, Jialong Tang, Jialin Wang, Jian Yang, Jianhong Tu, Jianwei Zhang, Jianxin Ma, Jin Xu, Jingren Zhou, Jinze Bai, Jinzheng He, Junyang Lin, Kai Dang, Keming Lu, Keqin Chen, Kexin Yang, Mei Li, Mingfeng Xue, Na~Ni, Pei Zhang, Peng Wang, Ru~Peng, Rui Men, Ruize Gao, Runji Lin, Shijie Wang, Shuai Bai, Sinan Tan, Tianhang Zhu, Tianhao Li, Tianyu Liu, Wenbin Ge, Xiaodong Deng, Xiaohuan Zhou, Xingzhang Ren, Xinyu Zhang, Xipin Wei, Xuancheng Ren, Yang Fan, Yang Yao, Yichang Zhang, Yu~Wan, Yunfei Chu, Yuqiong Liu, Zeyu Cui, Zhenru Zhang, and Zhihao Fan. 2024{\natexlab{a}}.
\newblock Qwen2 technical report.
\newblock \emph{arXiv preprint arXiv:2407.10671}.

\bibitem[{Yang et~al.(2024{\natexlab{b}})Yang, Li, Fang, and Chen}]{DBLP:conf/emnlp/YangLFC24}
Wanqi Yang, Yanda Li, Meng Fang, and Ling Chen. 2024{\natexlab{b}}.
\newblock Enhancing temporal sensitivity and reasoning for time-sensitive question answering.
\newblock In \emph{{EMNLP} (Findings)}, pages 14495--14508. Association for Computational Linguistics.

\bibitem[{Yuan et~al.(2024)Yuan, Xie, Huang, and Ananiadou}]{Back_to_future}
Chenhan Yuan, Qianqian Xie, Jimin Huang, and Sophia Ananiadou. 2024.
\newblock \href {https://doi.org/10.1145/3589334.3645376} {Back to the future: Towards explainable temporal reasoning with large language models}.
\newblock In \emph{Proceedings of the {ACM} on Web Conference 2024, {WWW} 2024, Singapore, May 13-17, 2024}, pages 1963--1974. {ACM}.

\bibitem[{Zhang et~al.(2020)Zhang, Kishore, Wu, Weinberger, and Artzi}]{BERTScore}
Tianyi Zhang, Varsha Kishore, Felix Wu, Kilian~Q. Weinberger, and Yoav Artzi. 2020.
\newblock Bertscore: Evaluating text generation with {BERT}.
\newblock In \emph{{ICLR}}. OpenReview.net.

\end{thebibliography}
\appendix

\section{Benchmark Details}
\label{sec:Benchmark Details}

\subsection{Prompt for Generating Explanations of Positive Samples}
\label{sec:Prompt for Generating Explanations of Positive Samples}

\begin{tcolorbox}[title = {Prompt for Positive Samples' Explanation}, center title]
\small
\begin{displayquote}
\emph{Given the following text: "we predict that [$e_s$] [$r$] [$e_o$] will happen on [$t_q$]. Here are the reasoning steps: $\mathcal{C}(e_s,e_o)$." Please revise the provided text to ensure that the prediction aligns with the reasoning steps. Expand the explanation of each reasoning step to make the text more coherent and readable. If necessary, add additional reasoning steps to clarify the logic. The output should be a single, concise paragraph without bullet points, ensuring clarity and logical consistency.}
\end{displayquote}
\end{tcolorbox}

\subsection{Prompt for Generating Explanations of Negative Samples}
\label{sec:Prompt for Generating Explanations of Negative Samples}

\begin{tcolorbox}[title = {Prompt for Negative Samples' Explanation}, center title]
\small
\begin{displayquote}
\emph{Given the following text: "It is plausible that [$e_s$] [$r$] [$e_o'$] will not happen on [$t_q$]. Here are the reasoning steps: $\mathcal{C}(e_s,e_o')$." Please revise the provided text to ensure that the prediction aligns with the reasoning steps. Expand the explanation of each reasoning step to make the text more coherent and readable. If necessary, add additional reasoning steps to clarify the logic. The output should be a single, concise paragraph without bullet points, ensuring clarity and logical consistency.}
\end{displayquote}
\end{tcolorbox}

\subsection{Prompt for Generating Explanations of Neutral Samples}
\label{sec:Prompt for Generating Explanations of Neutral Samples}

\begin{tcolorbox}[title = {Prompt for Neutral Samples' Explanation}, center title]
\small
\begin{displayquote}
\emph{Given the following text: "It is unsure that [$e_s$] [$r'$] [$e_o$] will happen on [$t_q$]. Here are the reasoning steps: $\mathcal{C}(e_s,e_o)$." Please revise the provided text to ensure that the prediction aligns with the reasoning steps. Expand the explanation of each reasoning step to make the text more coherent and readable. If necessary, add additional reasoning steps to clarify the logic. The output should be a single, concise paragraph without bullet points, ensuring clarity and logical consistency.}
\end{displayquote}
\end{tcolorbox}

\subsection{Example Prompt for Instruction Tuning}
\label{sec:Example Prompt for Instruction Tuning}

Here is an example of an instruction tuning prompt for the query: (Police (Australia), Make an appeal or request, Citizen (Australia), 2014-03-12).

\begin{tcolorbox}[title = {Example Prompt for Instruction Tuning}, center title]
\small
\begin{displayquote}
\emph{\textbf{Instruct:} Given the following document, is it plausible that Police (Australia) will Make an appeal or request Citizen (Australia) on 2014-03-12? Please answer yes, no, or unsure then explain your decision.} 

\emph{\textbf{Input:} Police (Australia) Appeal for aid Citizen (Australia) 2014-02-10. Police (Australia) Arrest, detain, or charge with legal action Citizen (Australia) 2014-02-25. Police (Australia) Arrest, detain, or charge with legal action Citizen (Australia) 2014-03-04. ...... Police (Australia) Arrest, detain, or charge with legal action Criminal (Australia) on 2014-02-27, Criminal (Australia) Sexually assault Citizen (Australia) on 2014-03-08......}
\end{displayquote}

\tcblower
\small
\begin{displayquote}
\emph{\textbf{Output:} Yes. Based on the information provided, it is plausible that the Police in Australia will make an appeal or request to citizens on March 12, 2014. This prediction is supported by a pattern of interactions between the police and citizens leading up to this date. The police have previously made appeals or requests to citizens on several occasions, including February 23, February 26, and March 11, 2014. Additionally, the police have been actively involved in various actions such as arrests, detentions, and investigations, indicating a heightened level of engagement with the public. The police have also been involved in more severe actions, such as using conventional military force and fighting with small arms, which suggests a context of ongoing tension or conflict. Given this backdrop, it is reasonable to anticipate that the police might continue their efforts to communicate with citizens, possibly to seek cooperation or provide updates on ongoing situations. Therefore, considering the frequency and nature of past interactions, it is likely that the police will make another appeal or request to citizens on March 12, 2014.}
\end{displayquote}
\end{tcolorbox}

\subsection{Benchmark Summary and Evaluation}
\label{sec:sec:Benchmark Summary and Evaluation}

 \begin{table}[!ht]
 \begin{center}
\resizebox{0.5\textwidth}{!}{
    \centering
    \resizebox{0.8\textwidth}{!}{
    \begin{tabular}{l|l|l|l|l|l|l}
    \hline
        Datasets & \#train & \#valid & \#test & \#entity & \#relations & time gap \\ \hline
        ICEWS14 & 74854 & 8514 & 7371 & 7128 & 230 & 1 day \\ \hline
        ICEWS18 & 373018 & 45995 & 49545 & 23033 & 256 & 1 day \\ \hline
        ICEWS05-15 & 368868 & 46302 & 46159 & 10094 & 251 & 1 day \\ \hline
        GDELT & 79319 & 9957 & 9715 & 5850 & 238 & 15 minutes \\ \hline
        WIKI & 539286 & 67538 & 63110 & 12554 & 24 & 1 year \\ \hline
    \end{tabular}
}}\end{center}
\caption{Dataset statistics.}
\label{tab: dataset}
\end{table}

The statistical details of the source data used to construct the benchmark are provided in Table~\ref{tab: dataset}. The data consist of three sources: the \textit{Integrated Crisis Early Warning System} (ICEWS), the \textit{Global Database of Events, Language, and Tone} (GDELT), and \textit{Wikipedia} (WIKI). Specifically, the ICEWS14 dataset includes events from 2014, the ICEWS18 dataset includes events from 2018, and the ICEWS05-15 dataset spans events from 2005 to 2015. The GDELT dataset records events at 15-minute intervals, while WIKI consists of Wikidata knowledge bases that store factual information with a time interval of one year. To ensure the quality and reliability of our dataset, we recruited three volunteers to evaluate the benchmark. Each volunteer assessed 200 randomly selected examples from the dataset. They were instructed to perform two key evaluations, assigning scores on a scale of 1 to 3 based on the following criteria:

\textbf{Explanation Text Quality (1-3):}
\begin{itemize}
    \item \textbf{1} - The explanation is unclear, incoherent, or unreasonable.
    \item \textbf{2} - The explanation is somewhat clear and reasonable but lacks coherence or completeness in certain aspects.
    \item \textbf{3} - The explanation is clear, coherent, and fully reasonable.
\end{itemize}

\textbf{Overall Consistency (1-3):}
\begin{itemize}
    \item \textbf{1} - The query text, reasoning chain, and explanation text are inconsistent or logically disconnected.
    \item \textbf{2} - There is partial consistency among the query text, reasoning chain, and explanation text, but logical gaps remain.
    \item \textbf{3} - The query text, reasoning chain, and explanation text are fully consistent and logically aligned.
\end{itemize}

The results of the human evaluation, as shown in Table~\ref{tab: volunteer_scores}, demonstrate a high level of accuracy and reliability in our benchmark generation process.

\begin{table}[!ht]
    \centering
    \resizebox{\linewidth}{!}{ 
    \begin{tabular}{l|l|l}
    \hline
        \textbf{Volunteer} & \textbf{Explanation Text Quality} & \textbf{Overall Consistency} \\ \hline
        Volunteer 1        & 2.80                             & 2.78                         \\ \hline
        Volunteer 2        & 2.74                             & 2.79                         \\ \hline
        Volunteer 3        & 2.86                             & 2.89                         \\ \hline
    \end{tabular}
    }
    \caption{Average scores for Explanation Text Quality and Overall Consistency by Volunteers.}
    \label{tab: volunteer_scores}
\end{table}

\section{Implementation Details}
\label{sec:Implementation Details}

\subsection{Baselines}
\label{sec:Baselines}

Below, we provide brief introductions to the LLMs used in our methods:
\begin{itemize}

    \item \textit{GPT-4o}~\cite{GPT-4o} is a large language model developed by OpenAI, representing an advanced iteration of the GPT series. It is known for its strong generalization capabilities across a wide range of natural language processing tasks, including reasoning, generation, and instruction-following.

    \item \textit{Llama-3.1-8B-Instruct}~\cite{Llama3-8b} is an instruction-tuned version of the Llama3 series, with 8 billion parameters.  The tuned versions use supervised fine-tuning (SFT) and reinforcement learning with human feedback (RLHF) to align with human preferences for helpfulness and safety. 
    
    \item \textit{Qwen2.5-7B-Instruct}~\cite{qwen2} is the latest series of Qwen large language models. It focuses on optimizing performance for instruction-based tasks.
    
    \item \textit{Mistral-7B-Instruct-v0.3}~\cite{mistral} is a 7-billion-parameter instruction-tuned model with an extended 32,768-token vocabulary, v3 tokenizer support, and function calling capabilities for improved task performance. 

\end{itemize}
We also introduce the graph-based methods (temporal encoders) utilized in our methods:
\begin{itemize}

    \item \textit{RE-GCN}~\cite{RE-GCN} proposes a recurrent evolution module based on relational GNNs to obtain embeddings that contain dynamic information for entities and relations.

    \item \textit{CEN}~\cite{CEN} uses a length-aware Convolutional Neural Network(CNN) to handle evolutional patterns of different lengths via an easy-to-difficult curriculum learning strategy.
    
     \item \textit{CENET}~\cite{CENET} aims to learn a robust distribution over the entire entity set and identify significant entities by leveraging both historical and non-historical dependencies within a contrastive learning framework.
         
    \item \textit{SiMFy}~\cite{SiMFy} is a straightforward method that combines MLP and historical frequency to model the temporal events.

\end{itemize}

\subsection{Hyperparameters}
\label{sec:Hyperparameters}

We set the window size $w$ to 30 and the threshold $\tau$ to 0.7 for constructing our benchmark. During training, the RE-GCN module is kept frozen, and LoRA is employed to fine-tune the model. The structural embedding size $d_{s}$ is set to 512, while the textual embedding size $d_{x}$ retains the original hidden layer dimensions of each LLM. The detailed hyperparameters used during training and inference are provided in Table~\ref{lora}. For optimization, we enable DeepSpeed ZeRO stage3~\footnote{\url{https://github.com/microsoft/Megatron-DeepSpeed}}. All models are trained and evaluated on 2 Nvidia A800 GPUs, each with 80GB of memory. 

\begin{table}[ht]
    \centering
    \resizebox{\linewidth}{!}{
    \begin{tabular}{c|c}
    \toprule
       \textbf{Name}  &  \textbf{Value} \\\midrule
        lora $r$ & 16 \\
        lora alpha & 32 \\
        lora dropout & 0.05 \\
        lora target modules & (q, k, v, o, down, up, gate) proj \\
        cutoff len & 2048 \\
        epochs & 3 \\
        per device batch size & 6 \\
        gradient accumulation steps & 1\\
        learning rate & $3e-4$ \\
        weight decay & $1e-5$ \\
        warm ratio & 0.01 \\
        lr scheduler type & cosine \\
        num return sequences & 10 \\
        projection layers & 1 \\\bottomrule
    \end{tabular}
    }
    \caption{Detailed hyperparameters used in our paper.}
    \label{lora}
\end{table}

\section{Additional Comparative Study Results}
\label{sec: Additional Comparative Study Results}

\subsection{Comparison with Graph-based Methods}
\label{Comparison with Graph-based Methods}

To provide a comprehensive comparison, we also evaluate four state-of-the-art graph-based methods(REGCN, CEN, CENET, and SiMFy) in comparison with our method on the task. Specifically, for the query \( (e_s, r, e_o, t_q) \), we utilize an MLP to adapt to our task, as defined below:
$$
P = \mathbf{W}_{query} (e_s \parallel r \parallel e_o) 
$$
where $\parallel$ denotes the concatenation operation, $P \in \mathbb{R}^3$, $e_s \in \mathbb{R}^{1 \times d_s}$, $r \in \mathbb{R}^{1 \times d_s}$, and $e_o \in \mathbb{R}^{1 \times d_s}$. Here, $\mathbf{W}_{query} \in \mathbb{R}^{3 \times 3d_s}$ is a learnable weight matrix, and $d_s$ represents the embedding dimension. 

The prediction results are presented in Table~\ref{tab: main results on graph-based methods table1} through Table~\ref{tab: main results on graph-based methods table3}. We can observe that GETER significantly outperforms existing graph-based in terms of prediction results. Furthermore, our approach provides human-readable inference processes, ensuring greater interpretability. In contrast, the intrinsic property of these graph-based methods is that they are \textbf{black-box models}, inherently lacking explainability and unable to generate explanation text. The detailed results of the prediction experiments are summarized in Table~\ref{tab: detailed prediction result1 on graph-based models}.

\begin{table}[!t]\small
\renewcommand{\arraystretch}{1.3}
 \centering
 \setlength{\tabcolsep}{1.4mm}
 \resizebox{\linewidth}{!}{
 \begin{tabular}{lcccccccccccc}\toprule
    \multirow{2}{*}{\textbf{Model}} & \multicolumn{4}{c}{\textbf{ICEWS14}} & \multicolumn{4}{c}{\textbf{GDELT}}
    \\\cmidrule(lr){2-5}\cmidrule(lr){6-9}
             & Positive & Negative & Neutral & Overall  & Positive & Negative & Neutral & Overall\\\midrule \specialrule{0em}{1.5pt}{1.5pt}
    RE-GCN & 52.76  & 55.00 & 75.38 & 59.97 & 57.62 & 59.39 & 84.18 & 66.22 \\
    CEN & 61.25 & 53.01 & 76.24 & 62.79 & 60.29 & 61.42 & 86.93 & 68.71 \\
    CENT & 55.03 & 60.82 & 78.27 & 63.60 & 61.34 & 62.71 & 87.98 & 69.84 \\
    SiMFy & 53.40 & 63.03 & 78.92 & 63.90  & \textbf{63.30} & 60.91 & 88.23 & 70.06 \\\midrule
    \textbf{GETER} & \textbf{77.45} & \textbf{75.73} & \textbf{85.15} & \textbf{79.08} & 61.29 & \textbf{68.92} & \textbf{88.59} & \textbf{72.02} 
    \\\bottomrule
 \end{tabular}}
 \caption{F1 scores (\%) of different graph-based methods on ICEWS14 and GDELT datasets.}
 \label{tab: main results on graph-based methods table1}
\end{table}

\begin{table}[!t]\small
\renewcommand{\arraystretch}{1.3}
 \centering
 \setlength{\tabcolsep}{1.4mm}
 \resizebox{\linewidth}{!}{
 \begin{tabular}{lcccccccccccc}\toprule
    \multirow{2}{*}{\textbf{Model}} & \multicolumn{4}{c}{\textbf{ICEWS05-15}} & \multicolumn{4}{c}{\textbf{ICEWS18}}
    \\\cmidrule(lr){2-5}\cmidrule(lr){6-9}
             & Positive & Negative & Neutral & Overall  & Positive & Negative & Neutral & Overall\\\midrule \specialrule{0em}{1.5pt}{1.5pt}
    REGCN & 65.10 & 63.53 & 83.57 & 70.50 & 62.13 & 58.81 & 81.53 & 67.03 \\
    CENET & 67.29 & 64.17 & 89.44 & 73.36 & 61.99 & 64.45 & 84.90 & 69.90 \\
    CEN & 63.27 & 65.57 & 86.88 & 71.59 & 59.82 & 60.86 & 79.19 & 66.16 \\
    SiMFy & 67.61 & 66.95 & 89.14 & 74.29 & 60.88 & 62.61 & 82.34 & 68.10 \\\midrule
    \textbf{GETER} & \textbf{78.94} & \textbf{76.48} & \textbf{90.38} & \textbf{81.80} & \textbf{75.61} & \textbf{75.94} & \textbf{87.51} & \textbf{79.40}  
    \\\bottomrule
 \end{tabular}}
 \caption{F1 scores (\%) of different graph-based methods on ICEWS05-15 and ICEWS18 datasets.}
 \label{tab: main results on graph-based methods table2}
\end{table}

\begin{table}[!t]\footnotesize 
\renewcommand{\arraystretch}{1.3} 
 \centering
 \setlength{\tabcolsep}{2mm} 
 \resizebox{0.8\linewidth}{!}{ 
 \begin{tabular}{lcccc}\toprule
    \multirow{2}{*}{\textbf{Model}} & \multicolumn{4}{c}{\textbf{WIKI}}
    \\\cmidrule(lr){2-5}
             & Positive & Negative & Neutral & Overall\\\midrule
    REGCN & 75.27 & 70.38 & 77.65 & 74.59 \\
    CENET & 76.11 & 77.06 & 83.51 & 78.86 \\
    CEN & 74.36 & 76.04 & 82.25 & 77.49 \\
    SiMFy & 79.03 & 77.39 & 81.53 & 79.40 \\\midrule
    \textbf{GETER} & \textbf{99.28} & \textbf{94.49} & \textbf{96.19} & \textbf{96.81} 
    \\\bottomrule
 \end{tabular}}
 \caption{F1 scores (\%) of different graph-based methods on the WIKI dataset.}
 \label{tab: main results on graph-based methods table3}
\end{table}

\subsection{Complexity Analysis of GETER}
\label{Complexity Analysis of GETER}

\begin{figure}[!t]
    \centering
    \includegraphics[width=0.9\linewidth]{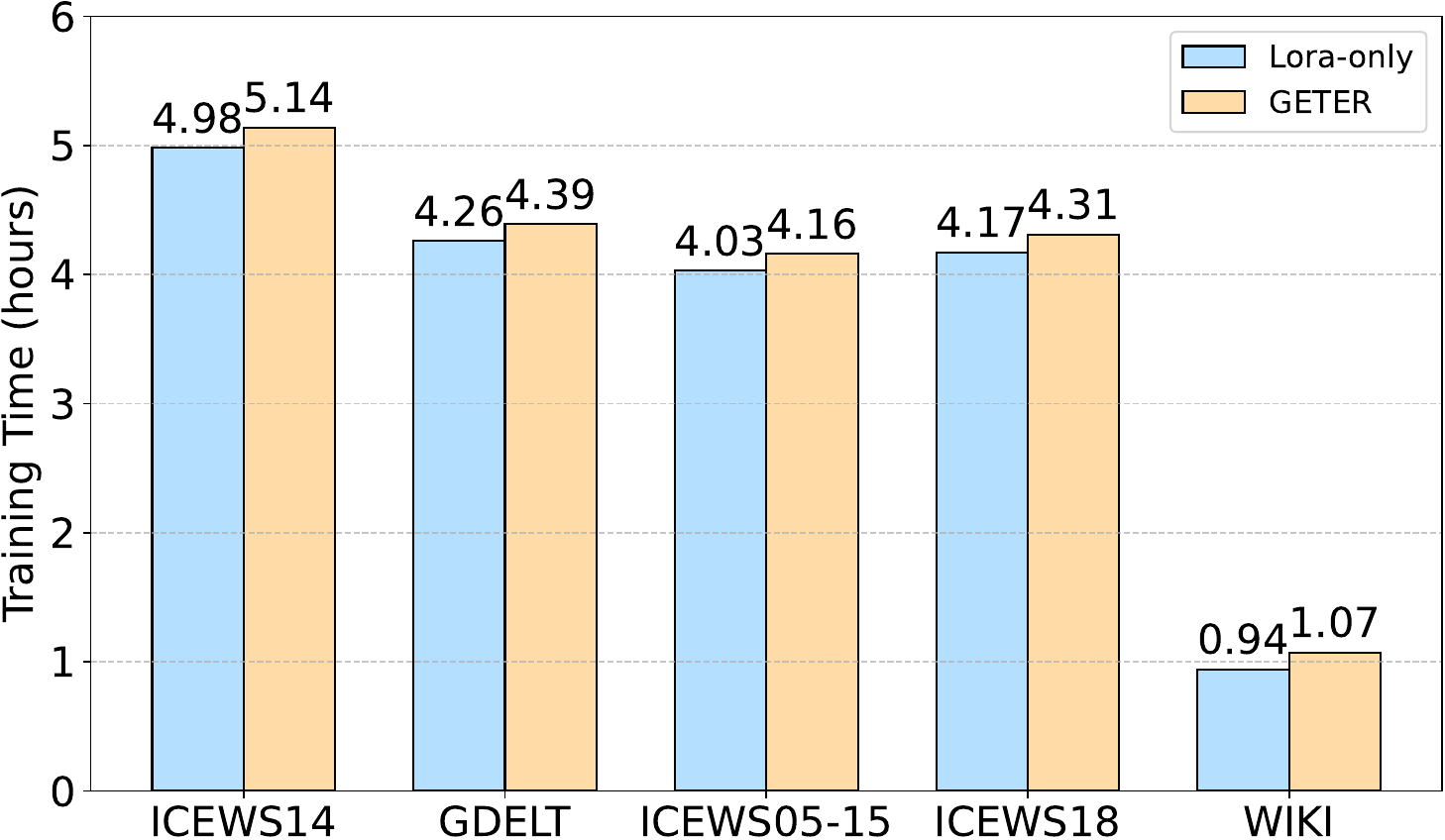}
    \caption{Comparison of training time between GETER and the LoRA fine-tuning method. The Y-axis represents the training time (hours).}
    \label{fig:Complexity of GETER}
\end{figure}

LLM applications often face challenges related to high computational costs due to the large number of model parameters. Specifically, for our method GETER, during the training and inference stages, the complexity  is $O(|L_1|^2 \cdot |L_2|)$ for the input-answer pair, where $|L_1|$ represents the length of the input text and $|L_2|$ represents the length of the answer. Considering the costs, we leverage Low-Rank Adaptation (LoRA) and DeepSpeed to accelerate both training and inference. Additionally, for a clearer comparison, we present the training time of GETER against LoRA fine-tuning methods across five datasets in Figure~\ref{fig:Complexity of GETER}. The results demonstrate that incorporating graph tokens into LLM fine-tuning introduces minimal additional time costs compared to simple LoRA fine-tuning. Furthermore, given the significant performance improvements achieved by our method, as detailed in Section~\ref{sec: Main results}, we consider these additional costs to be negligible.

\section{Case Study}
\label{sec:Case Study}

In this section, we present a case study to highlight the differences in responses among \textit{Inference-only} method, \textit{Tuned-only} method, and \textbf{GETER}. Specifically, we analyze the following positive query: \textit{(Police (Australia), Engage in material cooperation, Citizen (Australia), 2014-11-16)}, where the expected label is \textit{"Yes"}. As shown in Table~\ref{tab:case study}, \textit{Inference-only} method fail to capture the subtle cooperative signals embedded within the document (highlighted in orange), instead focusing primarily on dominant antagonistic actions, such as arrests and accusations, which result in incorrect negative predictions. While \textit{Tuned-only} method can observe cooperative signals and demonstrate an improved ability to incorporate the temporal aspects of events, they struggle to fully model the interplay between cooperative and antagonistic actions (highlighted in blue), leading to comparable negative predictions.  In contrast, \textbf{GETER} effectively captures the evolving patterns of event relationships and cooperative signals (highlighted in red). By leveraging explicit cues, such as requests and expressed intentions to cooperate, \textbf{GETER} not only predicts a positive outcome accurately but also provides the correct explanation.

\section{Full Experimental Results}
\label{sec:Full Experimental Results}

The prediction results for the ICEWS18 and WIKI datasets are summarized in Table~\ref{tab: Main results classification metrics on ICEWS18 and WIKI}, while the explanation results are detailed in Table~\ref{tab: Main results generate metrics on ICEWS18 and WIKI}. Notably, \textbf{GETER} demonstrates consistent and significant improvements across most metrics on these two datasets, underscoring its robustness and superior performance in complex scenarios. Compared to \textit{Tuned-only} methods, GETER combined with Mistral achieves overall F1 score improvements of $16.42\%$ and $10.35\%$ on the respective datasets. Additionally, the detailed prediction results for all five datasets are comprehensively summarized in Table~\ref{tab: prediction results on ICEWS14 dataset} through Table~\ref{tab: detailed prediction result1 on graph-based models}.

\clearpage
\begin{tcolorbox}[width=1\textwidth,title = { Query:  Police (Australia) will Engage in material cooperation Citizen (Australia) on 2014-11-16}, center title]
\small
\begin{displayquote}
\emph{\textbf{Instruction:} Given the following document, is it plausible that Police (Australia) will Engage in material cooperation Citizen (Australia) on 2014-11-16? Please answer yes, no, or unsure then explain your decision.}\\
\emph{\textbf{Input:} Police (Australia) Arrest, detain, or charge with legal action Citizen (Australia) 2014-11-12. Police (Australia) Arrest, detain, or charge with legal action Citizen (Australia) 2014-11-13. Police (Australia) Arrest, detain, or charge with legal action Citizen (Australia) 2014-11-14. Police (Australia) Accuse Citizen (Australia) 2014-11-02. Police (Australia) Accuse Citizen (Australia) 2014-11-09. Police (Australia) Accuse Citizen (Australia) 2014-11-10. Police (Australia) Make an appeal or request Citizen (Australia) 2014-11-12. \textcolor{orange}{Police (Australia) Make an appeal or request Citizen (Australia) 2014-11-13. Police (Australia) Make an appeal or request Citizen (Australia) 2014-11-14}. Police (Australia) fight with small arms and light weapons Citizen (Australia) 2014-11-09. Police (Australia) fight with small arms and light weapons Citizen (Australia) 2014-11-12. Police (Australia) fight with small arms and light weapons Citizen (Australia) 2014-11-14. Police (Australia) Use conventional military force Citizen (Australia) 2014-10-22. Police (Australia) Use conventional military force Citizen (Australia) 2014-10-24. Police (Australia) Use conventional military force Citizen (Australia) 2014-11-10. Police (Australia) Investigate Citizen (Australia) 2014-11-03. Police (Australia) Investigate Citizen (Australia) 2014-11-04. Police (Australia) Investigate Citizen (Australia) 2014-11-09. Police (Australia) Express intent to meet or negotiate Citizen (Australia) 2014-10-24. \textcolor{orange}{Police (Australia) Express intent to meet or negotiate Citizen (Australia) 2014-11-11. Police (Australia) Express intent to meet or negotiate Citizen (Australia) 2014-11-14}. Police (Australia) Criticize or denounce Citizen (Australia) 2014-10-28. Police (Australia) Criticize or denounce Citizen (Australia) 2014-11-03. Police (Australia) Confiscate property Citizen (Australia) 2014-10-30. Police (Australia) Investigate human rights abuses Citizen (Australia) 2014-10-30. Police (Australia) Appeal for intelligence Citizen (Australia) 2014-11-04. Police (Australia) Reject Citizen (Australia) 2014-11-07. Police (Australia) Abduct, hijack, or take hostage Citizen (Australia) 2014-11-09. Police (Australia) Physically assault Citizen (Australia) 2014-11-13. Police (Australia) Impose restrictions on political freedoms Citizen (Australia) 2014-11-14. Police (Australia) Return, release person(s) Citizen (Australia) 2014-11-14. Police (Australia) Arrest, detain, or charge with legal action Criminal (Australia) on 2014-10-23, Criminal (Australia) Engage in symbolic act Citizen (Australia) on 2014-11-03. Police (Australia) Arrest, detain, or charge with legal action Criminal (Australia) on 2014-10-23, Criminal (Australia) Sexually assault Citizen (Australia) on 2014-11-05. Police (Australia) Accuse Criminal (Australia) on 2014-10-22, Criminal (Australia) Engage in symbolic act Citizen (Australia) on 2014-11-03. Police (Australia) Accuse Criminal (Australia) on 2014-10-22, Criminal (Australia) Sexually assault Citizen (Australia) on 2014-11-05. Police (Australia) Demand Lawyer/Attorney (Australia) on 2014-10-31, Lawyer/Attorney (Australia) Arrest, detain, or charge with legal action Citizen (Australia) on 2014-11-03. Police (Australia) Demand Lawyer/Attorney (Australia) on 2014-10-31, Lawyer/Attorney (Australia) Accuse Citizen (Australia) on 2014-11-10. Police (Australia) Demand Lawyer/Attorney (Australia) on 2014-10-31, Lawyer/Attorney (Australia) Threaten Citizen (Australia) on 2014-11-05. Police (Australia) Confiscate property Criminal (Australia) on 2014-10-23, Criminal (Australia) Engage in symbolic act Citizen (Australia) on 2014-11-03. Police (Australia) Confiscate property Criminal (Australia) on 2014-10-23, Criminal (Australia) Sexually assault Citizen (Australia) on 2014-11-05.}\\
\emph{\textbf{Label output:} Yes. Based on the information provided, it is plausible that the Police in Australia will engage in material cooperation with Australian citizens on November 16, 2014. This prediction is supported by a series of interactions between the police and citizens leading up to this date. \textcolor{red} {The police have been actively involved in various legal actions, such as arresting, detaining, or charging citizens on multiple occasions from November 12 to November 14, 2014. Additionally, they have made appeals or requests to citizens during the same period, indicating a willingness to communicate and potentially collaborate. Despite instances of conflict, such as accusations, criticism, and even physical confrontations, there are also signs of negotiation and intent to meet, as evidenced by the police expressing their intent to negotiate on November 11 and November 14, 2014. Furthermore, the police have returned or released individuals on November 14, suggesting a move towards resolution and cooperation.} These actions, combined with the broader context of police investigations and appeals for intelligence, suggest a complex relationship that could lead to material cooperation on November 16, 2014.}
\end{displayquote}

\end{tcolorbox}

\begin{table*}[ht]
    \centering
    \footnotesize 
    \renewcommand{\arraystretch}{1.1} 
    \setlength{\tabcolsep}{3pt} 
    \resizebox{\textwidth}{!}{ 
    \begin{tabular}{l|l|p{0.9\textwidth}} 
    \toprule
    \multicolumn{3}{c}{\textbf{Positive Query:} Police (Australia) will Engage in material cooperation Citizen (Australia) on 2014-11-16  \textbf{ Model:} Mistral} \\ 
    \midrule
        \textbf{Methods} & \textbf{Labels} & \textbf{Output} \\
    \midrule
        \textit{Inference-only}  & \xmark & Based on the provided document, it is not plausible that Police (Australia) will engage in material cooperation with the Citizen (Australia) on 2014-11-16. The document shows a series of actions by the Police (Australia) that are generally antagonistic towards the Citizen (Australia), including arrests, detentions, charges, accusations, confiscation of property, physical assault, and investigations into human rights abuses. These actions suggest a conflictual relationship rather than cooperation. Furthermore, \textcolor{orange}{there is no indication in the document of any intention by the Police (Australia) to engage in cooperation with the Citizen (Australia)} on the specific date of 2014-11-16. \\
    \midrule
        \textit{Tuned-only} & \xmark & No. Based on the information provided, it is predicted that Police (Australia) will not engage in material cooperation with Citizen (Australia) on 2014-11-16. This prediction is supported by a series of events leading up to this date, which indicate a pattern of conflict and adversarial interactions between the police and citizens. \textcolor{blue}{In the days preceding 2014-11-16, the police have been involved in multiple actions against citizens, including arrests, detentions, and charges on 2014-11-12, 13, and 14. Additionally, there have been accusations against citizens on 2014-11-2, 9, and 10, as well as instances of police using force and military actions against citizens on various dates. The police have also been involved in investigations, criticisms, and confiscations, further indicating a tense relationship. Despite some expressions of intent to negotiate or meet, the overall pattern of behavior suggests a lack of cooperative engagement}. Given this context of ongoing conflict and legal actions, it is reasonable to conclude that material cooperation between the police and citizens is unlikely to occur on 2014-11-16. \\
    \midrule
        \textit{GETER (ours)} & \cmark & Yes. Based on the sequence of events outlined in the document, it is plausible to predict that the Police in Australia will engage in material cooperation with Australian citizens on November 16, 2014. This prediction is supported by a series of interactions between the police and citizens leading up to this date. From November 12 to November 14, the police were actively arresting, detaining, or charging citizens with legal actions, indicating a period of heightened enforcement. \textcolor{red}{\textbf{However, alongside these actions, the police also made several appeals and requests to citizens on the same dates, suggesting a willingness to engage in dialogue or seek assistance. Additionally, the police expressed an intent to meet or negotiate with citizens on multiple occasions, including November 11 and November 14, which implies a shift towards more collaborative engagement.}} Despite the confrontational actions, such as the use of force and accusations, the police also demonstrated a capacity for reconciliation by returning or releasing individuals on November 14. This dual approach of enforcement and negotiation, combined with the police's expressed intent to meet and cooperate, supports the likelihood of material cooperation occurring on November 16, 2014. \\
    \bottomrule
    \end{tabular}
    }
    \caption{Case comparisons between \textbf{GETER} and other methods. While \textit{Tuned-only} method demonstrate an improved ability to handle the temporal aspects of events (highlighted in blue), they still resulting in negative predictions. In contrast, \textbf{GETER} leverages temporal graph structures to model the evolving patterns of event relationships and effectively identifies cooperative signals (highlighted in red), enabling more accurate predictions.}
    \label{tab:case study}
\end{table*}

\begin{table*}[ht]
\begin{center}
\resizebox{1.0\textwidth}{!}{
\begin{tabular}{c|c|cccc|cccc} 
\toprule
\multirow{2}{*}{Models} & \multirow{2}{*}{\diagbox{Types}{Datasets}} & \multicolumn{4}{c|}{ICEWS18} & \multicolumn{4}{c}{WIKI} \\ \cline{3-10}
 &  & Positive & Negative & Neutral & Overall & Positive & Negative & Neutral & Overall \\ \midrule
\multirow{2}{*}{GPT-4o} 
 & zero-shot\textit{\ w/o chains text} & 51.64 & 36.61 & 24.79 & 38.32 &69.5 &53.45	&17.77 &47.45  \\
 & zero-shot & 60.33 & 23.78 & 40.72 & 42.08 &61.94	&37.44	&40.88	&47.54  \\ 
\midrule
\multirow{4}{*}{Llama3-8B-Instruct} 
 & zero-shot\textit{\ w/o chains text} & 7.68 & 24.39 & 38.95 & 22.93 &48.31	&54.39	&66.46	&52.44  \\
 & zero-shot & 55.12 & 18.81 & 9.14 & 28.79 &51.76	&26.43	&1.26	&27.31  \\
 & LoRA \textit{\ w/o chains text} & 57.47 & 47.14 & 56.30 & 53.66 &84.08	&70.67	&83.36	&79.80  \\
 & LoRA & 62.30 & 46.24 & 66.46 & 58.23 &88.59	&73.29	&81.36	&81.57  \\ 
 & \textbf{GETER} & \textbf{75.78} & \textbf{74.09} & \textbf{87.53} & \textbf{78.85} & \textbf{98.99} &\textbf{90.58}	&\textbf{91.00}	&\textbf{93.79}  \\ 
 & \cellcolor[gray]{0.9}$\Delta$Improve & \cellcolor[gray]{0.9}21.64\% & \cellcolor[gray]{0.9}60.24\% & \cellcolor[gray]{0.9}31.70\%  & \cellcolor[gray]{0.9}35.41\%  & \cellcolor[gray]{0.9}11.74\% & \cellcolor[gray]{0.9}23.59\% & \cellcolor[gray]{0.9}11.85\% & \cellcolor[gray]{0.9}14.98\% \\
\midrule
\multirow{4}{*}{Qwen2.5-7B-Instruct} 
 & zero-shot\textit{\ w/o chains text} &30.94  &40.53  &25.13  &32.34 &43.51	&53.31	&7.72	&34.54  \\
 & zero-shot &44.22   &48.67   &10.92   &35.40  &46.46	&47.84	&2.47	&32.23  \\
 & LoRA \textit{\ w/o chains text} &45.82   &59.83   &66.27   &56.82  &87.16	&80.29	&87.00	&85.04  \\
 & LoRA &69.68   &60.54   &63.21   &64.48  &88.65	&78.58	&87.36	&85.19  \\ 
 & \textbf{GETER} &\textbf{74.77} &\textbf{74.41} &\textbf{86.79} &\textbf{78.37}  &\textbf{97.32} &\textbf{93.33}&\textbf{94.01}	&\textbf{95.02}  \\ 
 & \cellcolor[gray]{0.9}$\Delta$Improve & \cellcolor[gray]{0.9}7.31\% &\cellcolor[gray]{0.9}22.91\%  & \cellcolor[gray]{0.9}37.28\% & \cellcolor[gray]{0.9}21.55\%  & \cellcolor[gray]{0.9}9.78\% & \cellcolor[gray]{0.9}18.77\% & \cellcolor[gray]{0.9}7.61\% & \cellcolor[gray]{0.9}11.54\% \\ 
\midrule
\multirow{4}{*}{Mistral-7B-Instruct} 
 & zero-shot\textit{\ w/o chains text} &1.06    &34.23   &47.64   &26.53 &35.81	&49.71	&55.40	&46.52  \\
 & zero-shot &4.14  &33.06  &41.58  &25.37  &62.98	&44.44	&41.89	&50.37  \\
 & LoRA \textit{\ w/o chains text} &58.07   &55.27   &74.46   &62.21  &84.94	&77.82	&83.08	&82.18  \\
 & LoRA &64.22   &64.63   &76.63   &68.20   &89.29	&86.61	&87.04	&87.73  \\ 
 & \textbf{GETER} &\textbf{75.61} &\textbf{75.94} &\textbf{87.51} &\textbf{79.40} &\textbf{99.28}	&\textbf{94.49}	&\textbf{96.19}	&\textbf{96.81}  \\ 
 & \cellcolor[gray]{0.9}$\Delta$Improve & \cellcolor[gray]{0.9}17.74\%  & \cellcolor[gray]{0.9}17.50\%  & \cellcolor[gray]{0.9}14.20\%  & \cellcolor[gray]{0.9}16.42\%  & \cellcolor[gray]{0.9}11.19\% & \cellcolor[gray]{0.9}9.10\% & \cellcolor[gray]{0.9}10.51\% & \cellcolor[gray]{0.9}10.35\% \\
\bottomrule
\end{tabular}
}
\end{center}
\caption{F1 scores (\%) of each model on the ICEWS18 and WIKI test instances. "Overall" represents the weighted average F1 score. \textit{w/o chains text} refers to the absence of reasoning chain input for LLMs. The best-performing results are highlighted in \textbf{bold}. $\Delta$Improve represents the relative improvements of \textbf{GETER} compared to \textbf{Tuned-only} methods.}
\label{tab: Main results classification metrics on ICEWS18 and WIKI}
\end{table*}

\begin{table*}[htb]
\begin{center}
\resizebox{1.0\textwidth}{!}{
\begin{tabular}{c|c|cccc|cccc} 
\toprule
\multirow{2}{*}{Models} & \multirow{2}{*}{\diagbox{Types}{Datasets}} & \multicolumn{4}{c|}{ICEWS18} & \multicolumn{4}{c}{WIKI} \\ \cline{3-10}
 &  & BLEU-4 & rougeL & METEOR & BertScore (F1) & BLEU-4 & rougeL & METEOR & BertScore (F1) \\ \midrule
\multirow{2}{*}{GPT-4o} 
 & zero-shot\textit{\ w/o chains text}&9.33   &22.67   &29.87   &67.48  &13.25	&28.18	&36.65	&69.10  \\
 & zero-shot &14.84  &31.16   &37.47   &72.98 &25.98	&41.77	&45.52	&78.69  \\ 
\midrule
\multirow{4}{*}{Llama3-8B-Instruct} 
 & zero-shot\textit{\ w/o chains text} &4.10   &15.85   &16.20   &61.14  &9.39	&25.41	&27.95	&66.88  \\
 & zero-shot &10.01  &29.52   &27.19   &70.01  &14.67	&36.67	&33.43	&75.85  \\
 & LoRA \textit{\ w/o chains text} &23.55  &35.95   &42.54   &78.02  &48.99	&63.53	&63.08	&87.13  \\
 & LoRA &37.33  &49.18   &53.05   &83.58 &52.09	&65.27	&66.67	&87.99  \\ 
 & \textbf{GETER} &\textbf{40.39}  &\textbf{52.12}   &\textbf{54.85}   &\textbf{84.60}  &\textbf{55.52}	&\textbf{68.06}	&\textbf{69.16}	&\textbf{88.77}  \\ 
 & \cellcolor[gray]{0.9}$\Delta$Improve & \cellcolor[gray]{0.9}8.20\% & \cellcolor[gray]{0.9}5.98\% & \cellcolor[gray]{0.9}3.39\% & \cellcolor[gray]{0.9}1.22\% & \cellcolor[gray]{0.9}6.59\% & \cellcolor[gray]{0.9}4.28\% & \cellcolor[gray]{0.9}3.73\% & \cellcolor[gray]{0.9}0.89\% \\ 
\midrule
\multirow{4}{*}{Qwen2.5-7B-Instruct} 
 & zero-shot\textit{\ w/o chains text} &7.02   &19.52   &30.09   &65.92  &7.33	&21.46	&34.56	&66.38  \\
 & zero-shot &10.46  &27.97   &26.80   &71.77 &20.21	&36.19	&41.52	&77.61  \\
 & LoRA \textit{\ w/o chains text} &25.50  &37.61   &42.91   &78.56  &51.84	&65.09	&65.59	&87.79  \\
 & LoRA &37.49  &49.61   &52.07   &83.60  &53.57	&67.19	&67.23	&88.59  \\ 
 & \textbf{GETER} &\textbf{38.99}  &\textbf{50.70}   &\textbf{53.79}   &\textbf{84.17} &\textbf{55.0} &\textbf{67.49}	&\textbf{70.00}	&\textbf{88.99}  \\ 
 & \cellcolor[gray]{0.9}$\Delta$Improve & \cellcolor[gray]{0.9}4.00\% & \cellcolor[gray]{0.9}2.20\% & \cellcolor[gray]{0.9}3.30\%  & \cellcolor[gray]{0.9}0.68\% & \cellcolor[gray]{0.9}2.67\% & \cellcolor[gray]{0.9}0.45\% & \cellcolor[gray]{0.9}4.12\% & \cellcolor[gray]{0.9}0.45\% \\ 
\midrule
\multirow{4}{*}{Mistral-7B-Instruct} 
 & zero-shot\textit{\ w/o chains text} &7.60   &19.43   &23.98   &65.87  &11.41	&26.23	&31.64	&67.79  \\
 & zero-shot &9.74   &29.00   &26.00   &71.95  &21.25	&40.05	&41.43	&77.27  \\
 & LoRA \textit{\ w/o chains text} &25.46  &37.62   &42.91   &78.67  &51.58	&66.32	&65.29	&87.96  \\
 & LoRA &36.96  &49.12   &51.70   &83.38  &52.61	&65.40	&66.80	&87.97  \\ 
 & \textbf{GETER} &\textbf{39.64}  &\textbf{51.62}   &\textbf{54.04}   &\textbf{84.37}  &\textbf{54.96}	&\textbf{67.74}	&\textbf{69.17}	&\textbf{88.92}  \\ 
 & \cellcolor[gray]{0.9}$\Delta$Improve & \cellcolor[gray]{0.9}7.25\% & \cellcolor[gray]{0.9}5.09\% & \cellcolor[gray]{0.9}4.53\% & \cellcolor[gray]{0.9}1.19\% & \cellcolor[gray]{0.9}4.47\% & \cellcolor[gray]{0.9}3.58\% & \cellcolor[gray]{0.9}3.55\% & \cellcolor[gray]{0.9}1.08\% \\ 
\bottomrule
\end{tabular}
}
\end{center}
\caption{The semantic similarity performance (\%) of each model on the ICEWS18 and WIKI test instances. \textit{w/o chains text} refers to the absence of reasoning chain input for LLMs. The best-performing results are highlighted in \textbf{bold}.}
\label{tab: Main results generate metrics on ICEWS18 and WIKI}
\end{table*}


\begin{table*}[ht]
\begin{center}
\resizebox{1.0\textwidth}{!}{
\begin{tabular}{c|c|ccc|ccc|ccc|ccc} 
\toprule
\multirow{2}{*}{Models} & \multirow{2}{*}{Types} & \multicolumn{3}{c|}{Positive} & \multicolumn{3}{c|}{Negative} & \multicolumn{3}{c|}{Neutral} & \multicolumn{3}{c}{Overall} \\ \cline{3-14}
 &  & Precision & Recall & F1 & Precision & Recall & F1 & Precision & Recall & F1 & Precision & Recall & F1 \\ \midrule
\multirow{2}{*}{GPT-4o} 
 & zero-shot \textit{w/o chains text} &41.89  &72.62  &53.13  &24.63  &16.86  &20.02  &23.08  &9.00  &12.95  &30.76  &35.86  &30.61    \\
 & zero-shot &58.53  &61.75  &60.10  &33.05  &5.57   &9.54   &37.08  &70.33 &48.56  &43.91  &45.48  &39.95  \\ \midrule
\multirow{4}{*}{Llama3-8B-Instruct} 
 & zero-shot \textit{w/o chains text} &41.94  &14.62  &21.69  &30.23  &24.57  &27.11  &26.20  &54.67 &35.42  &33.54  &29.38  &27.42  \\
 & zero-shot &40.59  &93.00  &56.51  &21.17  &6.71   &10.20  &44.44  &3.33  &6.20   &38.62  &35.22  &26.70  \\
 & LoRA \textit{w/o chains text} &54.16  &73.25  &62.27  &35.28  &38.86  &36.98  &82.59  &34.00 &48.17  &55.99  &50.57  &49.81  \\
 & LoRA &66.08  &75.25  &70.37  &56.75  &59.43  &58.06  &78.73  &59.83 &67.99  &66.59  &65.57  &65.59  \\ 
& \cellcolor[gray]{0.9}\textbf{GETER} & \cellcolor[gray]{0.9}71.62 & \cellcolor[gray]{0.9}78.87 & \cellcolor[gray]{0.9}75.07 & \cellcolor[gray]{0.9}66.90 & \cellcolor[gray]{0.9}67.86 & \cellcolor[gray]{0.9}67.38 & \cellcolor[gray]{0.9}88.41 & \cellcolor[gray]{0.9}75.00 & \cellcolor[gray]{0.9}81.15 & \cellcolor[gray]{0.9}74.85 & \cellcolor[gray]{0.9}74.10 & \cellcolor[gray]{0.9}74.25 \\\midrule
\multirow{4}{*}{Qwen2.5-7B-Instruct} 
 & zero-shot \textit{w/o chains text} &41.51  &16.50  &23.61  &32.18  &62.71  &42.54  &17.94  &12.50 &14.73  &31.67  &30.76  &27.39  \\
 & zero-shot &55.79  &50.62  &53.08  &34.79  &65.00  &45.32  &57.58  &6.33  &11.41  &49.30  &42.76  &38.59  \\
 & LoRA \textit{w/o chains text} &66.85  &59.25  &62.82  &51.23  &68.43  &58.59  &83.33  &63.33 &71.97  &66.35  &63.48  &64.03  \\
 & LoRA &74.32  &74.88  &74.60  &61.58  &70.29  &65.64  &83.64  &69.00 &75.62  &72.73  &71.67  &71.90  \\ 
& \cellcolor[gray]{0.9}\textbf{GETER} & \cellcolor[gray]{0.9}81.56 & \cellcolor[gray]{0.9}71.88 & \cellcolor[gray]{0.9}76.41 & \cellcolor[gray]{0.9}68.84 & \cellcolor[gray]{0.9}81.43 & \cellcolor[gray]{0.9}74.61 & \cellcolor[gray]{0.9}86.95 & \cellcolor[gray]{0.9}82.17 & \cellcolor[gray]{0.9}84.49 & \cellcolor[gray]{0.9}78.86 & \cellcolor[gray]{0.9}78.00 & \cellcolor[gray]{0.9}78.12 \\\midrule
\multirow{4}{*}{Mistral-7B-Instruct} 
 & zero-shot \textit{w/o chains text} &68.18  &1.87   &3.65   &38.38  &40.57  &39.44  &33.63  &75.00 &46.44  &48.38  &35.67  &27.81  \\
 & zero-shot &55.56  &13.75  &22.04  &27.43  &27.86  &27.64  &30.65  &60.83 &40.76  &39.06  &31.90  &29.26  \\
 & LoRA \textit{w/o chains text} &77.89  &46.25  &58.04  &56.13  &78.43  &65.44  &77.13  &83.17 &80.03  &70.42  &67.52  &66.79  \\
 & LoRA &72.56  &73.38  &72.96  &60.66  &73.57  &66.49  &87.56  &64.50 &74.28  &72.88  &70.90  &71.18  \\ 
& \cellcolor[gray]{0.9}\textbf{GETER} & \cellcolor[gray]{0.9}83.62 & \cellcolor[gray]{0.9}72.12 & \cellcolor[gray]{0.9}77.45 & \cellcolor[gray]{0.9}69.23 & \cellcolor[gray]{0.9}83.57 & \cellcolor[gray]{0.9}75.73 & \cellcolor[gray]{0.9}87.79 & \cellcolor[gray]{0.9}82.67 & \cellcolor[gray]{0.9}85.15 & \cellcolor[gray]{0.9}80.02 & \cellcolor[gray]{0.9}78.95 & \cellcolor[gray]{0.9}79.08 \\
\bottomrule
\end{tabular}
}
\end{center}
\caption{Precision (\%), Recall (\%), and F1 scores (\%) for each model on the ICEWS14 dataset.}
\label{tab: prediction results on ICEWS14 dataset}
\end{table*}

\begin{table*}[ht]
\begin{center}
\resizebox{1.0\textwidth}{!}{
\begin{tabular}{c|c|ccc|ccc|ccc|ccc} 
\toprule
\multirow{2}{*}{Models} & \multirow{2}{*}{Types} & \multicolumn{3}{c|}{Positive} & \multicolumn{3}{c|}{Negative} & \multicolumn{3}{c|}{Neutral} & \multicolumn{3}{c}{Overall} \\ \cline{3-14}
  &  & Precision & Recall & F1 & Precision & Recall & F1 & Precision & Recall & F1 & Precision & Recall & F1 \\ \midrule
\multirow{2}{*}{GPT-4o} 
 & zero-shot \textit{w/o chains text} &44.70  &12.12  &19.08  &32.97  &65.14  &43.78  &27.82  &23.54  &25.50  &35.78  &32.84  &29.06   \\
 & zero-shot &49.67  &37.50  &42.74  &32.48  &43.43  &37.16  &30.16  &28.31  &29.21  &38.18  &36.65  &36.83   \\ \midrule
\multirow{4}{*}{Llama3-8B-Instruct} 
 & zero-shot \textit{w/o chains text} &38.10  &1.00   &1.95   &31.80  &34.57  &33.13  &29.09  &61.23  &39.44  &33.32  &30.14  &23.44   \\
 & zero-shot  &41.04  &76.75  &53.48  &26.92  &11.00  &15.62  &40.76  &23.08  &29.47  &36.36  &39.12  &33.90   \\
 & LoRA \textit{w/o chains text} &97.75  &45.33  &61.94  &3.86   &52.94  &7.19   &54.46  &94.65  &69.14  &54.09  &62.72  &46.29    \\
 & LoRA &81.87  &51.01  &62.86  &19.14  &56.30  &28.57  &77.23  &79.94  &78.56  &60.05  &61.48  &56.44    \\ 
& \cellcolor[gray]{0.9}\textbf{GETER} & \cellcolor[gray]{0.9}75.49 & \cellcolor[gray]{0.9}53.50 & \cellcolor[gray]{0.9}62.62 & \cellcolor[gray]{0.9}59.10 & \cellcolor[gray]{0.9}82.14 & \cellcolor[gray]{0.9}68.74 & \cellcolor[gray]{0.9}91.64 & \cellcolor[gray]{0.9}86.00 & \cellcolor[gray]{0.9}88.73 & \cellcolor[gray]{0.9}75.03 & \cellcolor[gray]{0.9}72.65 & \cellcolor[gray]{0.9}72.51 \\\midrule
\multirow{4}{*}{Qwen2.5-7B-Instruct} 
 & zero-shot \textit{w/o chains text} &34.18  &6.75   &11.27  &32.27  &73.86  &44.92  &26.41  &15.85  &19.81  &31.21  &31.35  &24.81   \\
 & zero-shot &57.89  &13.75  &22.22  &32.77  &91.29  &48.23  &40.00  &0.62   &1.21   &44.30  &35.02  &24.34   \\
 & LoRA \textit{w/o chains text} &22.88  &49.46  &31.28  &90.71  &36.56  &52.11  &6.62   &100.00 &12.41  &40.05  &60.54  &32.36   \\
 & LoRA &13.13  &76.09  &22.39  &92.71  &40.74  &56.61  &54.92  &85.20  &66.79  &51.67  &67.33  &46.95  \\ 
& \cellcolor[gray]{0.9}\textbf{GETER} & \cellcolor[gray]{0.9}75.64 & \cellcolor[gray]{0.9}55.13 & \cellcolor[gray]{0.9}63.77 & \cellcolor[gray]{0.9}61.40 & \cellcolor[gray]{0.9}81.57 & \cellcolor[gray]{0.9}70.06 & \cellcolor[gray]{0.9}89.32 & \cellcolor[gray]{0.9}87.54 & \cellcolor[gray]{0.9}88.42 & \cellcolor[gray]{0.9}75.14 & \cellcolor[gray]{0.9}73.53 & \cellcolor[gray]{0.9}73.27 \\\midrule 
\multirow{4}{*}{Mistral-7B-Instruct} 
 & zero-shot \textit{w/o chains text} &34.29  &3.00   &5.52   &30.40  &61.29  &40.64  &23.17  &23.85  &23.50  &29.66  &28.28  &22.39   \\
 & zero-shot &44.44  &0.50   &0.99   &34.34  &19.57  &24.93  &29.68  &79.54  &43.23  &36.69  &30.60  &21.55   \\
 & LoRA \textit{w/o chains text} &11.00  &83.81  &19.45  &97.71  &41.40  &58.16  &57.38  &94.91  &71.52  &53.26  &73.36  &47.80   \\
 & LoRA &64.88  &56.78  &60.56  &55.29  &54.89  &55.09  &73.85  &90.40  &81.29  &64.47  &66.33  &65.05  \\ 
& \cellcolor[gray]{0.9}\textbf{GETER} & \cellcolor[gray]{0.9}76.21 & \cellcolor[gray]{0.9}51.25 & \cellcolor[gray]{0.9}61.29 & \cellcolor[gray]{0.9}58.15 & \cellcolor[gray]{0.9}84.57 & \cellcolor[gray]{0.9}68.92 & \cellcolor[gray]{0.9}92.76 & \cellcolor[gray]{0.9}84.77 & \cellcolor[gray]{0.9}88.59 & \cellcolor[gray]{0.9}75.33 & \cellcolor[gray]{0.9}72.23 & \cellcolor[gray]{0.9}72.02 \\
\bottomrule
\end{tabular}
}
\end{center}
\caption{Precision (\%), Recall (\%), and F1 scores (\%) for each model on the GDELT dataset.}
\label{tab: prediction results on GDELT dataset}
\end{table*}

\begin{table*}[ht]
\begin{center}
\resizebox{1.0\textwidth}{!}{
\begin{tabular}{c|c|ccc|ccc|ccc|ccc} 
\toprule
\multirow{2}{*}{Models} & \multirow{2}{*}{Types} & \multicolumn{3}{c|}{Positive} & \multicolumn{3}{c|}{Negative} & \multicolumn{3}{c|}{Neutral} & \multicolumn{3}{c}{Overall} \\ \cline{3-14}
  &  & Precision & Recall & F1 & Precision & Recall & F1 & Precision & Recall & F1 & Precision & Recall & F1 \\ \midrule
\multirow{2}{*}{GPT-4o} 
 & zero-shot \textit{w/o chains text} &46.72  &68.19  &55.45  &25.52  &27.21  &26.33  &25.70  &11.06  &15.47  &32.99  &36.36  &33.03   \\
 & zero-shot &65.32  &58.33  &61.63  &22.45  &8.09   &11.89  &36.86  &65.45  &47.16  &42.05  &44.03  &40.58   \\ \midrule
\multirow{4}{*}{Llama3-8B-Instruct} 
 & zero-shot \textit{w/o chains text} &42.98  &6.81   &11.75  &31.60  &26.76  &28.98  &29.20  &60.61  &39.41  &34.81  &30.63  &26.30  \\
 & zero-shot &41.62  &91.11  &57.14  &25.28  &13.38  &17.50  &44.35  &8.33   &14.03  &37.10  &38.93  &30.24  \\
 & LoRA \textit{w/o chains text} &51.29  &91.25  &65.67  &51.73  &30.74  &38.56  &93.87  &53.33  &68.02  &65.08  &59.13  &57.47  \\
 & LoRA &74.92  &68.06  &71.32  &70.56  &40.88  &51.77  &61.46  &94.24  &74.40  &69.17  &67.48  &65.86   \\ 
& \cellcolor[gray]{0.9}\textbf{GETER} & \cellcolor[gray]{0.9}73.37 & \cellcolor[gray]{0.9}84.58 & \cellcolor[gray]{0.9}78.58 & \cellcolor[gray]{0.9}83.84 & \cellcolor[gray]{0.9}69.41 & \cellcolor[gray]{0.9}75.95 & \cellcolor[gray]{0.9}91.00 & \cellcolor[gray]{0.9}91.97 & \cellcolor[gray]{0.9}91.48 & \cellcolor[gray]{0.9}82.48 & \cellcolor[gray]{0.9}81.94 & \cellcolor[gray]{0.9}81.84 \\\midrule
\multirow{4}{*}{Qwen2.5-7B-Instruct} 
 & zero-shot \textit{w/o chains text} &46.63  &24.03  &31.71  &30.80  &54.85  &39.45  &18.83  &13.64  &15.82  &32.50  &30.87  &29.17  \\
 & zero-shot &60.88  &30.69  &40.81  &33.99  &83.53  &48.32  &23.08  &0.91   &1.75   &39.89  &38.59  &30.78   \\
 & LoRA \textit{w/o chains text} &79.27  &42.50  &55.33  &56.24  &88.09  &68.65  &89.49  &82.58  &85.89  &74.95  &70.39  &69.52   \\
 & LoRA &84.18  &55.42  &66.83  &62.07  &82.79  &70.95  &82.92  &85.30  &84.09  &76.48  &74.03  &73.72  \\ 
& \cellcolor[gray]{0.9}\textbf{GETER} & \cellcolor[gray]{0.9}71.68 & \cellcolor[gray]{0.9}86.11 & \cellcolor[gray]{0.9}78.23 & \cellcolor[gray]{0.9}84.36 & \cellcolor[gray]{0.9}64.26 & \cellcolor[gray]{0.9}72.95 & \cellcolor[gray]{0.9}88.77 & \cellcolor[gray]{0.9}91.06 & \cellcolor[gray]{0.9}89.90 & \cellcolor[gray]{0.9}81.34 & \cellcolor[gray]{0.9}80.49 & \cellcolor[gray]{0.9}80.23 \\\midrule
\multirow{4}{*}{Mistral-7B-Instruct} 
 & zero-shot \textit{w/o chains text} &61.70  &4.03   &7.56   &34.78  &31.76  &33.21  &33.98  &71.67  &46.10  &43.93  &34.85  &28.37  \\
 & zero-shot &59.52  &10.42  &17.73  &35.97  &25.44  &29.80  &36.13  &79.55  &49.69  &44.25  &37.52  &31.96  \\
 & LoRA \textit{w/o chains text} &66.30  &75.97  &70.81  &77.39  &26.18  &39.12  &62.79  &95.61  &75.80  &68.84  &65.83  &61.95  \\
 & LoRA &77.44  &68.19  &72.53  &68.77  &75.44  &71.95  &82.94  &85.45  &84.18  &76.34  &76.12  &76.07  \\ 
& \cellcolor[gray]{0.9}\textbf{GETER} & \cellcolor[gray]{0.9}75.67 & \cellcolor[gray]{0.9}82.50 & \cellcolor[gray]{0.9}78.94 & \cellcolor[gray]{0.9}82.85 & \cellcolor[gray]{0.9}71.03 & \cellcolor[gray]{0.9}76.48 & \cellcolor[gray]{0.9}88.29 & \cellcolor[gray]{0.9}92.58 & \cellcolor[gray]{0.9}90.38 & \cellcolor[gray]{0.9}82.08 & \cellcolor[gray]{0.9}81.94 & \cellcolor[gray]{0.9}81.80 \\
\bottomrule
\end{tabular}
}
\end{center}
\caption{Precision (\%), Recall (\%), and F1 scores (\%) for each model on the ICEWS05-15 dataset.}
\label{tab: prediction results on ICEWS05-15 dataset}
\end{table*}

\begin{table*}[ht]
\begin{center}
\resizebox{1.0\textwidth}{!}{
\begin{tabular}{c|c|ccc|ccc|ccc|ccc} 
\toprule
\multirow{2}{*}{Models} & \multirow{2}{*}{Types} & \multicolumn{3}{c|}{Positive} & \multicolumn{3}{c|}{Negative} & \multicolumn{3}{c|}{Neutral} & \multicolumn{3}{c}{Overall} \\ \cline{3-14}
&  & Precision & Recall & F1 & Precision & Recall & F1 & Precision & Recall & F1 & Precision & Recall & F1 \\ \midrule
\multirow{2}{*}{GPT-4o} 
 & zero-shot \textit{w/o chains text} & 47.43 & 56.67 & 51.64 & 38.20 & 35.14 & 36.61 & 26.79 & 23.08 & 24.79 & 37.96 & 39.10 & 38.32 \\
 & zero-shot & 53.09 & 69.87 & 60.33 & 34.51 & 18.14 & 23.78 & 38.12 & 43.69 & 40.72 & 42.26 & 44.52 & 42.08 \\ \midrule
\multirow{4}{*}{Llama3-8B-Instruct} 
 & zero-shot \textit{w/o chains text} & 25.19 & 4.53 & 7.68 & 31.25 & 20.00 & 24.39 & 27.82 & 64.92 & 38.95 & 28.02 & 28.38 & 22.93 \\
 & zero-shot & 39.37 & 91.87 & 55.12 & 33.09 & 13.14 & 18.81 & 45.83 & 5.08 & 9.14 & 39.28 & 38.76 & 28.79 \\
 & LoRA \textit{w/o chains text} & 57.20 & 57.73 & 57.47 & 39.31 & 58.86 & 47.14 & 90.17 & 40.92 & 56.30 & 61.44 & 52.90 & 53.66 \\
 & LoRA & 75.14 & 53.20 & 62.30 & 49.43 & 43.43 & 46.24 & 55.87 & 82.00 & 66.46 & 60.61 & 58.86 & 58.23 \\ 
& \cellcolor[gray]{0.9}\textbf{GETER} & \cellcolor[gray]{0.9}72.42 & \cellcolor[gray]{0.9}79.47 & \cellcolor[gray]{0.9}75.78 & \cellcolor[gray]{0.9}78.87 & \cellcolor[gray]{0.9}69.86 & \cellcolor[gray]{0.9}74.09 & \cellcolor[gray]{0.9}87.06 & \cellcolor[gray]{0.9}88.00 & \cellcolor[gray]{0.9}87.53 & \cellcolor[gray]{0.9}79.10 & \cellcolor[gray]{0.9}78.90 & \cellcolor[gray]{0.9}78.85 \\ \midrule
\multirow{4}{*}{Qwen2.5-7B-Instruct} 
 & zero-shot \textit{w/o chains text} & 43.98 & 23.87 & 30.94 & 34.15 & 49.86 & 40.53 & 24.74 & 25.54 & 25.13 & 34.75 & 33.05 & 32.34 \\
 & zero-shot & 58.01 & 35.73 & 44.22 & 35.55 & 77.14 & 48.67 & 35.29 & 6.46 & 10.92 & 43.49 & 40.48 & 35.40 \\
 & LoRA \textit{w/o chains text} & 72.00 & 33.60 & 45.82 & 45.02 & 89.14 & 59.83 & 92.31 & 51.69 & 66.27 & 69.29 & 57.71 & 56.82 \\
 & LoRA & 69.50 & 69.87 & 69.68 & 52.46 & 71.57 & 60.54 & 84.14 & 50.62 & 63.21 & 64.63 & 68.35 & 64.48 \\ 
& \cellcolor[gray]{0.9}\textbf{GETER} & \cellcolor[gray]{0.9}74.87 & \cellcolor[gray]{0.9}74.67 & \cellcolor[gray]{0.9}74.77 & \cellcolor[gray]{0.9}72.88 & \cellcolor[gray]{0.9}76.00 & \cellcolor[gray]{0.9}74.41 & \cellcolor[gray]{0.9}88.75 & \cellcolor[gray]{0.9}84.92 & \cellcolor[gray]{0.9}86.79 & \cellcolor[gray]{0.9}78.50 & \cellcolor[gray]{0.9}78.29 & \cellcolor[gray]{0.9}78.37 \\ \midrule
\multirow{4}{*}{Mistral-7B-Instruct} 
 & zero-shot \textit{w/o chains text} & 57.14 & 0.53 & 1.06 & 35.71 & 32.86 & 34.23 & 34.51 & 76.92 & 47.64 & 42.99 & 34.95 & 26.53 \\
 & zero-shot & 69.57 & 2.13 & 4.14 & 35.04 & 31.29 & 33.06 & 30.10 & 67.23 & 41.58 & 45.84 & 32.00 & 25.37 \\
 & LoRA \textit{w/o chains text} & 69.52 & 49.87 & 58.07 & 62.68 & 49.43 & 55.27 & 61.19 & 95.08 & 74.46 & 64.66 & 63.71 & 62.21 \\
 & LoRA & 73.41 & 57.07 & 64.22 & 62.80 & 66.57 & 64.63 & 70.45 & 84.00 & 76.63 & 68.96 & 68.57 & 68.20 \\ 
& \cellcolor[gray]{0.9}\textbf{GETER} & \cellcolor[gray]{0.9}75.36 & \cellcolor[gray]{0.9}75.87 & \cellcolor[gray]{0.9}75.61 & \cellcolor[gray]{0.9}73.98 & \cellcolor[gray]{0.9}78.00 & \cellcolor[gray]{0.9}75.94 & \cellcolor[gray]{0.9}90.61 & \cellcolor[gray]{0.9}84.62 & \cellcolor[gray]{0.9}87.51 & \cellcolor[gray]{0.9}79.62 & \cellcolor[gray]{0.9}79.29 & \cellcolor[gray]{0.9}79.40 \\
\bottomrule
\end{tabular}
}
\end{center}
\caption{Precision (\%), Recall (\%), and F1 scores (\%) for each model on the ICEWS18 dataset.}
\label{tab: prediction results on ICEWS18 dataset}
\end{table*}

\begin{table*}[ht]
\begin{center}
\resizebox{1.0\textwidth}{!}{
\begin{tabular}{c|c|ccc|ccc|ccc|ccc} 
\toprule
\multirow{2}{*}{Models} & \multirow{2}{*}{Types} & \multicolumn{3}{c|}{Positive} & \multicolumn{3}{c|}{Negative} & \multicolumn{3}{c|}{Neutral} & \multicolumn{3}{c}{Overall} \\ \cline{3-14}
&  & Precision & Recall & F1 & Precision & Recall & F1 & Precision & Recall & F1 & Precision & Recall & F1 \\ \midrule
\multirow{2}{*}{GPT-4o} 
 & zero-shot \textit{w/o chains text} &66.93	&72.33	&69.53	&40.85	&77.27	&53.45	&93.94	&9.81	&17.77	&68.07	&53.00	&47.45  \\
 & zero-shot &52.51	&75.50	&61.94	&33.80	&41.96	&37.44	&88.42	&26.58	&40.88	&58.83	&49.10	&47.54  \\ \midrule
\multirow{4}{*}{Llama3-8B-Instruct} 
 & zero-shot \textit{w/o chains text} &68.98	&37.18	&48.31	&51.74	&57.34	&54.39	&47.19	&66.46	&66.46	&56.53	&53.00	&52.44  \\
 & zero-shot &39.23	&76.08	&51.76	&27.01	&25.87	&26.43	&100.00	&0.63	&1.26	&55.78	&35.83	&27.31  \\
 & LoRA \textit{w/o chains text} &91.35	&77.89	&84.08	&67.83	&73.76	&70.67	&78.48	&88.89	&83.36	&79.98	&80.31	&79.80\\
 & LoRA  &96.25&82.06	&88.59	&70.98	&75.75	&73.29	&75.95	&87.59	&81.36	&81.88	&82.00	&81.57  \\ 
& \cellcolor[gray]{0.9}\textbf{GETER} & \cellcolor[gray]{0.9}99.13 & \cellcolor[gray]{0.9}98.85 & \cellcolor[gray]{0.9}98.99 & \cellcolor[gray]{0.9}85.89 & \cellcolor[gray]{0.9}95.80 & \cellcolor[gray]{0.9}90.58 & \cellcolor[gray]{0.9}96.13 & \cellcolor[gray]{0.9}86.39 & \cellcolor[gray]{0.9}91.00 & \cellcolor[gray]{0.9}94.14 & \cellcolor[gray]{0.9}93.78 & \cellcolor[gray]{0.9}93.79 \\\midrule
\multirow{4}{*}{Qwen2.5-7B-Instruct} 
 & zero-shot \textit{w/o chains text} &52.44	&37.18	&43.51	&37.83	&90.21	&53.31	&61.90	&4.11	&7.72	&51.19	&42.15	&34.54  \\
 & zero-shot &73.29	&34.01	&46.46	&32.69	&89.16	&47.84	&50.00	&1.27	&2.47	&53.30	&39.73	&32.23  \\
 & LoRA \textit{w/o chains text} &91.93	&82.86	&87.16	&76.22	&84.82	&80.29	&85.76	&88.27	&87.00	&85.14	&85.25	&85.04  \\
 & LoRA &83.29	&94.75	&88.65	&85.31	&72.84	&78.58	&86.39	&88.35	&87.36	&84.93	&86.02	&85.19  \\ 
& \cellcolor[gray]{0.9}\textbf{GETER} & \cellcolor[gray]{0.9}95.30 & \cellcolor[gray]{0.9}99.42 & \cellcolor[gray]{0.9}97.32 & \cellcolor[gray]{0.9}96.28 & \cellcolor[gray]{0.9}90.56 & \cellcolor[gray]{0.9}93.33 & \cellcolor[gray]{0.9}93.71 & \cellcolor[gray]{0.9}94.30 & \cellcolor[gray]{0.9}94.01 & \cellcolor[gray]{0.9}95.07 & \cellcolor[gray]{0.9}95.05 & \cellcolor[gray]{0.9}95.02 \\\midrule
\multirow{4}{*}{Mistral-7B-Instruct} 
 & zero-shot \textit{w/o chains text} &73.87	&23.63	&35.81	&54.85	&45.45	&49.71	&42.26	&80.38	&55.40	&57.62	&49.10	&46.52  \\
 & zero-shot &75.20	&54.18	&62.98	&38.95	&51.75	&44.44	&41.69	&42.09	&41.89	&53.12	&49.42	&50.37  \\
 & LoRA \textit{w/o chains text} &86.17	&83.75	&84.94	&72.38	&84.15	&77.82	&87.03	&79.48	&83.08	&82.30	&82.45	&82.18  \\
 & LoRA &93.66	&85.30	&89.29	&82.52	&91.12	&86.61	&86.08	&88.03	&87.04	&87.78	&87.96	&87.73  \\ 
& \cellcolor[gray]{0.9}\textbf{GETER} & \cellcolor[gray]{0.9}98.86 & \cellcolor[gray]{0.9}99.71 & \cellcolor[gray]{0.9}99.28 & \cellcolor[gray]{0.9}99.61 & \cellcolor[gray]{0.9}89.86 & \cellcolor[gray]{0.9}94.49 & \cellcolor[gray]{0.9}92.67 & \cellcolor[gray]{0.9}100.00 & \cellcolor[gray]{0.9}96.19 & \cellcolor[gray]{0.9}97.02 & \cellcolor[gray]{0.9}96.84 & \cellcolor[gray]{0.9}96.81 \\
\bottomrule
\end{tabular}
}
\end{center}
\caption{Precision (\%), Recall (\%), and F1 scores (\%) for each model on the WIKI dataset.}
\label{tab: prediction results on WIKI dataset}
\end{table*}


\begin{table*}[!t]
\begin{center}
\resizebox{1.0\textwidth}{!}{
\begin{tabular}{c|c|ccc|ccc|ccc|ccc} 
\toprule
\multirow{2}{*}{Datasets} & \multirow{2}{*}{Methods} & \multicolumn{3}{c|}{Positive} & \multicolumn{3}{c|}{Negative} & \multicolumn{3}{c|}{Neutral} & \multicolumn{3}{c}{Overall} \\ \cline{3-14}
&  & Precision & Recall & F1 & Precision & Recall & F1 & Precision & Recall & F1 & Precision & Recall & F1 \\ \midrule
\multirow{4}{*}{ICEWS14} 
 & REGCN & 58.02 & 48.38 & 52.76 & 48.89 & 62.86 & 55.00 & 80.11 & 71.17 & 75.38 & 61.29 & 59.71 & 59.97 \\
 & CENET & 70.51 & 45.12 & 55.03 & 53.03 & 71.29 & 60.82 & 75.43 & 81.33 & 78.27 & 66.09 & 64.19 & 63.60 \\
 & CEN & 60.29 & 62.25 & 61.25 & 56.05 & 50.29 & 53.01 & 73.53 & 79.17 & 76.24 & 62.66 & 63.10 & 62.79 \\
 & SiMFy & 68.84 & 43.63 & 53.40 & 53.43 & 76.86 & 63.03 & 79.86 & 78.00 & 78.92 & 66.85 & 64.52 & 63.90 \\ 
 & \cellcolor[gray]{0.9}\textbf{GETER} & \cellcolor[gray]{0.9}83.62 & \cellcolor[gray]{0.9}72.12 & \cellcolor[gray]{0.9}77.45 & \cellcolor[gray]{0.9}69.23 & \cellcolor[gray]{0.9}83.57 & \cellcolor[gray]{0.9}75.73 & \cellcolor[gray]{0.9}87.79 & \cellcolor[gray]{0.9}82.67 & \cellcolor[gray]{0.9}85.15 & \cellcolor[gray]{0.9}80.02 & \cellcolor[gray]{0.9}78.95 & \cellcolor[gray]{0.9}79.08 \\
 \midrule
\multirow{4}{*}{GDELT} 
 & REGCN & 61.93 & 53.87 & 57.62 & 55.61 & 63.71 & 59.39 & 84.05 & 84.31 & 84.18 & 66.56 & 66.28 & 66.22 \\
 & CENET & 65.58 & 57.63 & 61.34 & 59.39 & 66.43 & 62.71 & 87.05 & 88.92 & 87.98 & 70.05 & 69.95 & 69.84 \\
 & CEN & 64.80 & 56.37 & 60.29 & 57.21 & 66.29 & 61.42 & 87.40 & 86.46 & 86.93 & 69.16 & 68.70 & 68.71 \\
 & SiMFy & 63.11 & 63.50 & 63.30 & 60.39 & 61.43 & 60.91 & 89.42 & 87.08 & 88.23 & 70.18 & 69.95 & 70.06 \\ 
 & \cellcolor[gray]{0.9}\textbf{GETER} & \cellcolor[gray]{0.9}76.21 & \cellcolor[gray]{0.9}51.25 & \cellcolor[gray]{0.9}61.29 & \cellcolor[gray]{0.9}58.15 & \cellcolor[gray]{0.9}84.57 & \cellcolor[gray]{0.9}68.92 & \cellcolor[gray]{0.9}92.76 & \cellcolor[gray]{0.9}84.77 & \cellcolor[gray]{0.9}88.59 & \cellcolor[gray]{0.9}75.33 & \cellcolor[gray]{0.9}72.23 & \cellcolor[gray]{0.9}72.02 \\
 \midrule
 \multirow{4}{*}{ICEWS05-15} 
 & REGCN & 66.96 & 63.33 & 65.10 & 65.57 & 61.62 & 63.53 & 79.05 & 88.64 & 83.57 & 70.38 & 70.87 & 70.50 \\
 & CENET & 65.27 & 69.44 & 67.29 & 65.30 & 63.09 & 64.17 & 91.05 & 87.88 & 89.44 & 73.54 & 73.25 & 73.36 \\
 & CEN & 67.45 & 59.58 & 63.27 & 61.33 & 70.44 & 65.57 & 88.02 & 85.76 & 86.88 & 72.02 & 71.55 & 71.59 \\
 & SiMFy & 69.03 & 66.25 & 67.61 & 65.45 & 68.53 & 66.95 & 89.35 & 88.94 & 89.14 & 74.36 & 74.27 & 74.29 \\ 
 & \cellcolor[gray]{0.9}\textbf{GETER} & \cellcolor[gray]{0.9}75.67 & \cellcolor[gray]{0.9}82.50 & \cellcolor[gray]{0.9}78.94 & \cellcolor[gray]{0.9}82.85 & \cellcolor[gray]{0.9}71.03 & \cellcolor[gray]{0.9}76.48 & \cellcolor[gray]{0.9}88.29 & \cellcolor[gray]{0.9}92.58 & \cellcolor[gray]{0.9}90.38 & \cellcolor[gray]{0.9}82.08 & \cellcolor[gray]{0.9}81.94 & \cellcolor[gray]{0.9}81.80 \\
 \midrule
 \multirow{4}{*}{ICEWS18} 
 & REGCN & 59.34 & 65.20 & 62.13 & 62.82 & 55.29 & 58.81 & 80.91 & 82.15 & 81.53 & 67.18 & 67.14 & 67.03 \\
 & CENET & 65.38 & 58.93 & 61.99 & 62.58 & 66.43 & 64.45 & 82.97 & 86.92 & 84.90 & 69.89 & 70.10 & 69.90 \\
 & CEN & 60.60 & 59.07 & 59.82 & 59.38 & 62.43 & 60.86 & 80.25 & 78.15 & 79.19 & 66.28 & 66.10 & 66.16 \\
 & SiMFy & 64.71 & 57.47 & 60.88 & 61.29 & 64.00 & 62.61 & 79.23 & 85.69 & 82.34 & 68.07 & 68.38 & 68.10 \\ 
 & \cellcolor[gray]{0.9}\textbf{GETER} & \cellcolor[gray]{0.9}75.36 & \cellcolor[gray]{0.9}75.87 & \cellcolor[gray]{0.9}75.61 & \cellcolor[gray]{0.9}73.98 & \cellcolor[gray]{0.9}78.00 & \cellcolor[gray]{0.9}75.94 & \cellcolor[gray]{0.9}90.61 & \cellcolor[gray]{0.9}84.62 & \cellcolor[gray]{0.9}87.51 & \cellcolor[gray]{0.9}79.62 & \cellcolor[gray]{0.9}79.29 & \cellcolor[gray]{0.9}79.40 \\
 \midrule
 \multirow{4}{*}{WIKI} 
 & REGCN & 81.76 & 69.74 & 75.27 & 72.32 & 68.53 & 70.38 & 70.94 & 85.76 & 77.65 & 75.31 & 74.71 & 74.59 \\
 & CENET & 77.95 & 74.35 & 76.11 & 77.19 & 76.92 & 77.06 & 81.38 & 85.76 & 83.51 & 78.86 & 78.93 & 78.86 \\
 & CEN & 77.50 & 71.47 & 74.36 & 78.44 & 73.78 & 76.04 & 77.22 & 87.97 & 82.25 & 77.69 & 77.66 & 77.49 \\
 & SiMFy & 78.25 & 79.83 & 79.03 & 82.03 & 73.43 & 77.39 & 78.76 & 84.49 & 81.53 & 79.56 & 79.45 & 79.40 \\ 
 & \cellcolor[gray]{0.9}\textbf{GETER} & \cellcolor[gray]{0.9}98.86 & \cellcolor[gray]{0.9}99.71 & \cellcolor[gray]{0.9}99.28 & \cellcolor[gray]{0.9}99.61 & \cellcolor[gray]{0.9}89.86 & \cellcolor[gray]{0.9}94.49 & \cellcolor[gray]{0.9}92.67 & \cellcolor[gray]{0.9}100.00 & \cellcolor[gray]{0.9}96.19 & \cellcolor[gray]{0.9}97.02 & \cellcolor[gray]{0.9}96.84 & \cellcolor[gray]{0.9}96.81 \\
\bottomrule
\end{tabular}
}
\end{center}
\caption{Precision (\%), Recall (\%), and F1 scores (\%) for each graph-based model across different datasets. "Overall" represents the weighted average F1 score.}
\label{tab: detailed prediction result1 on graph-based models}
\end{table*}

\end{document}